\documentclass[12pt]{article}
\usepackage{amsthm}
\usepackage{color}
\usepackage{makeidx}  
\usepackage{multirow}      
\usepackage{subfigure,graphicx}
\usepackage{epstopdf}
\usepackage{subfig}
\usepackage{amsmath,dsfont}
\usepackage{amsfonts}
\usepackage{graphicx,epstopdf}
\usepackage{enumitem}
\usepackage{color}
\usepackage{times}
\usepackage{bbding}
\usepackage{wasysym}
\usepackage[T1]{fontenc}
\usepackage{colonequals}
\usepackage{psfrag}
\usepackage{subfigure}
\usepackage{mathrsfs}
\usepackage{lscape}
\usepackage{cite}
\usepackage{setspace}
\usepackage{pifont,textcomp,bbm}
\usepackage{algorithm,algpseudocode}
\usepackage{dsfont}
\usepackage{amsmath,calc}
\usepackage{multicol}
\usepackage{booktabs}
\newcommand{\blind}{0}
\usepackage{booktabs}
\newtheorem{thm}{Theorem}
\newtheorem{lm}{Lemma}

\newtheorem*{proof*}{Proof}
 
\newcommand{\compresslist}{
  \vspace{-1em}
  \setlength{\itemsep}{1pt}
  \setlength{\parskip}{0pt}
  \setlength{\parsep}{0pt}
}

\begin{document}

\def\spacingset#1{\renewcommand{\baselinestretch}%
{#1}\small\normalsize} \spacingset{1}

\date{}
\if0\blind
{
  \title{\bf Or's of And's for Interpretable Classification, with Application to Context-Aware Recommender Systems}
  
  \author{Tong Wang\hspace{.2cm}\\
    Department of Electrical Engineering and Computer Science\\ Massachussetts Institute of Technology\\
    Cynthia Rudin \\
    Sloan School of Management\\ Massachussetts Institute of Technology\\
    Finale Doshi-Velez \\
    Department of Computer Science\\
    Harvard University\\
    Yimin Liu, Erica Klampfl and Perry MacNeille\\
    Ford Motor Company
    }
  \maketitle
} \fi


\bigskip
\begin{abstract}
We present a machine learning algorithm for building classifiers that are comprised of a small number of disjunctions of conjunctions (\textit{or}'s of \textit{and}'s). An example of a classifier of this form is as follows: If X satisfies ($x_1 = $ `blue' AND $x_3=$ `middle') OR ($x_1 = $ `blue' AND $x_2=$ `<15') OR ($x_1 = $ `yellow'), then we predict that Y=1, ELSE predict Y=0. An attribute-value pair is called a \emph{literal} and a conjunction of literals is called a \emph{pattern}. Models of this form have the advantage of being interpretable to human experts, since they produce a set of conditions that concisely describe a specific class. We present two probabilistic models for forming a pattern set, one with a Beta-Binomial prior, and the other with Poisson priors. In both cases, there are prior parameters that the user can set to encourage the model to have a desired size and shape, to conform with a domain-specific definition of interpretability. 
We provide two scalable MAP inference approaches: 
a pattern level search, which involves association rule mining, and a literal level search. We 
show stronger priors reduce computation.
We apply the Bayesian Or's of And's (\emph{BOA}) model to predict user behavior with respect to in-vehicle context-aware personalized recommender systems.  
\end{abstract}

\noindent%
{\it Reproducibility: }all code and datasets will be made publicly available on acceptance
\vfill

\newpage
\spacingset{1.45} 
\section{Introduction}
Our goal is to construct a model that not only classifies data but also ``explains'' data. This predictive classification model consists of a small number of disjunctions of conjunctions, that is, the classifiers are \textit{or}'s of \textit{and}'s, which have both the classification and expressive power. This form of model has some recent precedent in  \cite{FriedmanFi99,MalioutovVa13,hauser2010disjunctions,GohRu14} as a form of model that is natural for modeling consumer behavior, and 
interpretable to human experts. 
In particular, it has been hypothesized that people make purchasing decisions using simple \textit{or}'s of \textit{and}'s models (e.g., ``I would only purchase product X if it has options 1 and 2 or options 3 and 4."). The set of conditions within the model should be sparse, as humans can handle about 7$\pm$2 cognitive entities at once \cite{miller1956magical}. Beyond modeling human decision-making, \textit{or}'s of \textit{and}'s models can strike a nice balance between accuracy and interpretability for general predictive modeling problems.


We consider the classification problem where we observe $\{\mathbf{x}_n,y_n\}$ pairs, where $\mathbf{x}_n$ is a vector of real-valued, categorical, or binary attributes and $y_n\in \{0,1\}$. 
A \textit{literal} is an attribute-value pair (e.g., $x_1$=`blue'), 
denoted as $r$. 
A \textit{pattern} is a conjunctions of literals (e.g., $x_1$=`blue' AND $x_2$=`<5' AND $x_3$=`middle'), denoted as $a$. It has the form $a = r_1 \wedge r_2 \wedge ...$, where $\wedge$ denotes the \emph{and} operation. We call the number of literals in $a$ the \emph{length} of $a$. A \textit{pattern set} is a disjunction of patterns, denoted as $A$. It has the form $A =  a_1 \vee a_2 \vee ...$, where $\vee$ denotes the \textit{or} operation. We call the number of patterns in $A$ the \emph{size} of $A$.


We take a generative approach to the construction of \textit{or}'s of \textit{and}'s classifiers and introduce two models, a model with beta-binomial priors, called \text{BOA-BetaBinomial}, and a model with Poisson priors, called \text{BOA-Poisson}. Both models have priors that can be adjusted to suit a domain-specific notion of interpretability, as it is well-known that interpretability comes in different forms for different domains  \cite{martens2010building, martens2011performance, huysmans2011empirical, allahyari2011user, ruping2006learning,Freitas:2014ic}. In particular, the prior parameters of the BOA-Poisson are the expected \textit{pattern lengths} and \textit{pattern set size}. The parameters of BOA-BetaBinomial are pseudo counts for the number of conjunctions of each size to be selected within the model. That is, if the user desires conjunctions that have two conditions each, the pseudo-counts for size two conditions can be increased. Users can set these parameters to obtain a model of an approximate desired size, though it is possible for the prior to be overwhelmed with data. 

We provide two inference methods. The first technique uses a combination of association rule mining and simulated annealing to approximate the globally optima BOA maximum a posteriori (MAP) model. 
This approach is motivated by a theoretical bound that allows us to reduce the size of the computationally hard problem of finding the MAP solution so that it may be more manageable to solve in practice. This bound states that 
we need only mine patterns that are sufficiently frequent in the database. 
The second method uses a literal-based stochastic local search where a neighboring point is proposed by changing a literal in the current pattern set. Each method has its own advantages and disadvantages: the first method has the disadvantage that it potentially requires generating and screening a huge number of patterns, but it is more likely to find an MAP solution, and the final model will be comprised of only high quality, pre-screened patterns. The second inference method does not pre-screen the patterns, so it searches over a much larger space in theory, at the expense of a more difficult computation. In practice either method can be used, though if the user has a preference for patterns of shorter lengths, the first method could be substantially faster and lead to better solutions.

Our applied interest is to understand user response to personalized
advertisements that are chosen based on the user, the advertisement,
and the context.  Such systems are called \emph{context-aware
  recommender systems} (see surveys 
\cite{Adomavicius2005,Adomavicius08,verbert2012context,baldauf2007survey} and references therein). One major challenge in the design of recommender systems,
reviewed in \cite{verbert2012context}, is the \textit{interaction
  challenge}: users typically wish to know \emph{why} certain
recommendations were made. Our work addresses precisely this challenge: our models provide patterns in data that describe conditions on which a recommendation will be accepted. 
\section{Related Work}
The models we are studying have different names in different fields: ``disjunctions of conjunctions" in marketing, ``classification patterns" in data mining, and ``disjunctive normal forms" (DNF) in artificial intelligence. Learning logical models of this form has an extensive history. Valiant \cite{valiant84} showed that DNFs could be learned in polynomial time in the PAC (probably approximately correct) setting, and recent work has improved those bounds via polynomial threshold functions~\cite{klivans01} and Fourier analysis~\cite{feldman12}.  However, these theoretical approaches often require unrealistic modeling assumptions and do not incorporate a user-control over interpretability.

In parallel, the data-mining literature has developed approaches to
building logical conjunctive models.  Associative classification methods (e.g., 
\cite{ma1998integrating, li2001cmar,yin2003cpar,chen2006new,cheng2007discriminative,McCormickRuMa12,RudinLeMa13}) mine for frequent patterns in the data and combine them to build classifiers, generally in a heuristic way, where patterns are ranked by an interestingness criteria and the top several patterns are used. Some of these methods, like CBA, CPAR and CMAR \cite{li2001cmar, yin2003cpar, chen2006new, cheng2007discriminative} still suffer from a huge number of patterns and do not yield interpretable classifiers, yet it is well-known that for many domains, the space of good predictive models is often large enough to include very simple models \cite{holte93}. Inductive logic programming \cite{muggleton1994inductive} is similar, in that it mines (potentially complicated) patterns and takes the simple union of these patterns as the pattern set, rather than optimizing the pattern set directly. This is a major disadvantage over the approach we take here. 
Another class of approaches aim to construct DNF models by greedily adding the conjunction that explains the most of the remaining data \cite{MalioutovVa13, pollack1988pediatric, FriedmanFi99, gaines1995induction, cohen1995fast}. Thus, again, these methods do not directly aim to produce globally optimal conjunctive models.  There are few recent techniques that do aim to fully learn DNF models \cite{hauser2010disjunctions,GohRu14}, which present integer programming approaches for solving the full problems, and also present relaxations for computational efficiency. These are very different from our work in that the method of Hauser \emph{et al.}\cite{hauser2010disjunctions} is not generative and also does not have the advantage of reduction to a smaller problem that we have. The work of Goh and Rudin \cite{GohRu14} is for real valued features only, whereas we focus mainly on categorical data, though binned or thresholded real-valued data would suffice where the bins need not be exclusive.

Note that logical models are generally robust to outliers and naturally handle missing data, with no imputation needed for missing attribute values. These methods can perform comparably with traditional convex optimization-based methods such as support vector machines or lasso (though linear models are not always considered to be interpretable in many domains).   

\textit{Or}'s of \textit{and}'s models are also a special case of another form of interpretable model called $M$-of-$N$ rules\cite{Freitas:2014ic, towell1993extracting, chevaleyre2013rounding, ordonez1999testing, UstunRu14}, inparticular when $M$=1. In an $M$-of-$N$ rules model, an example is classified as positive if at least $M$ criteria among $N$ are satisfied. If $M$=1, the model becomes a disjunction of conditions, and if $M$=$N$, then the model is a single conjunction. (In these models, one rule generally refers to one literal, whereas in our model, each pattern can have multiple literals.)

The main application we consider is in-vehicle context-aware recommender systems. The most similar works to ours include that of Baralis \emph{et al.}\cite{baralis2011cas}, who present a framework that discovers relationships between user context and services using association rules. Lee at al. \cite{lee2006location} create interpretable context-aware recommendations by using a decision tree model that considers location context, personal context, environmental context and user preferences. However,  they did not study some of the most important factors we include, namely contextual information such as the user's destination, relative locations of services along the route, the location of the services with respect to the user's destination,  passenger(s) in the vehicle, etc. Our work is related to recommendation systems for in-vehicle context-aware music recommendations, see \cite{baltrunas2011incarmusic,WangRoWa12}, but whether a user will accept a music recommendation does not depend on anything analogous to the location of a business that the user would drive to.
 The setup of in-vehicle recommendation systems are also different than, for instance, mobile-tourism guides\cite{Noguera201237,schwinger2005context,van2004context,tung2004personalized} where the user is searching to accept a recommendation, and interacts heavily with the system in order to find an acceptable recommendation. The closest work to ours is probably that of Park \emph{et al}.\cite{park2007location} who also consider Bayesian predictive models for context aware recommender systems to restaurants. They also consider demographic and context-based attributes. They did not study advertising, however, which means they did not consider the locations to the coupon's venue, expiration times, etc.

\section{Bayesian or's of and's}\label{sec:boa}
We work with standard classification data. The data set
$S$ consists of $\{\mathbf{x}_n,y_n\}_{n=1,...N}$, where $y_n\in \{0,1\}$ and $\{\mathbf{x}_n\}_{n=1,...N}$ has $N$ observations and $J$ attributes. 
$S^+$ is the class of observations with positive labels, and the
observations with negative labels are $S^-$.  We use $A$ to represent a set of patterns and a pattern in $A$ is represented as $a_i$, indexed by $i \in\{1,...,|A|\}$. We define a boolean function $h(\mathbf{x}_n,a)$ that evaluates if pattern $a$ applies to data point $\mathbf{x}_n$. Then we can define a classifier built from $A$ as $f_A$.
\begin{equation}
f_A(\mathbf{x}_n) = \begin{cases} 1 & \exists a \in A, h(\mathbf{x}_n,a)=1 \\  0 & \text{otherwise.} \end{cases}
\end{equation}
As long as a data point satisfies at least one of the patterns in $A$, it is classified as positive.

\begin{figure}[h]
\centering
\includegraphics[width=0.45\textwidth,height=0.3\textwidth]{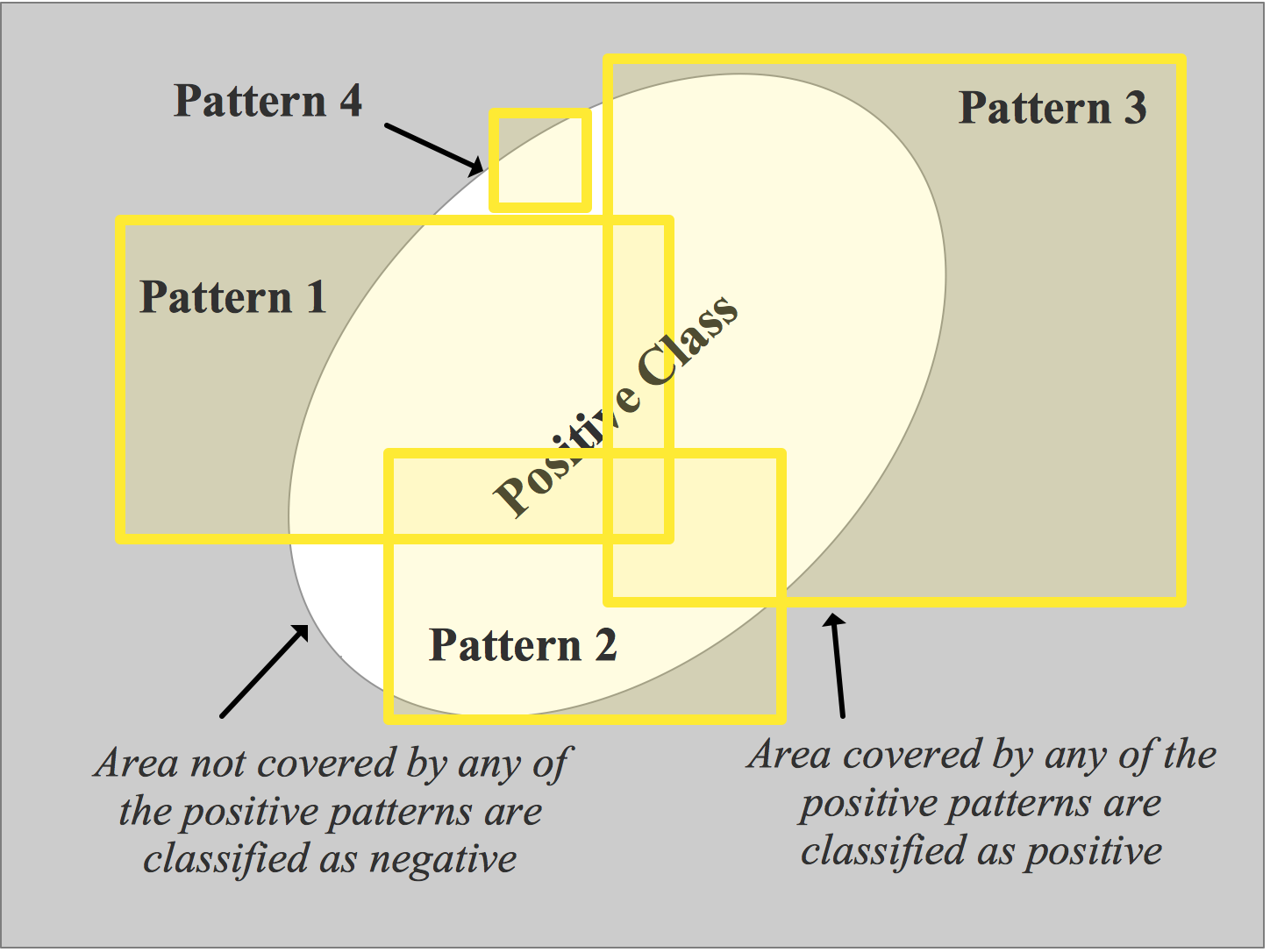}
\caption{Illustration of or's of and's}
\label{fig:DDR}
\end{figure}

Figure~\ref{fig:DDR} shows an example of a pattern set. Each pattern is a yellow patch that covers a particular area, and the pattern applies to the area it covers. In Figure~\ref{fig:DDR}, the white oval in the middle indicates the positive class. Our goal is to find a set of patterns $A$ that covers mostly the positive class, but little of the negative class. 

We present a probabilistic model for selecting patterns.  Taking a Bayesian approach allows us to flexibly incorporate users' expectations on the ``shape" of a pattern set through the prior.  The user can guide the model toward more domain-interpretable solutions by specifying a desired balance between the size and lengths of patterns.

\subsection{Prior}
We propose two models for the prior. In the BOA-BetaBinomial model, the maximum length $L$ of patterns is pre-determined by a user. Patterns of the same length are placed into the same pattern \textit{pool}. The model uses $L$ beta priors to control the probabilities of selecting patterns from different pools. In a BOA-Poisson model, the ``shape'' of a pattern set, which includes the number of patterns and lengths of patterns, is decided by drawing from Poisson distributions parameterized with user-defined values. Then the generative process fills it in with literals by first randomly selecting attributes and then randomly selecting values corresponding to each attribute. We present the two prior models in detail.
\subsubsection{Beta-Binomial prior}\label{sec:betabinomial}
Notice that the pattern set we want to find should include patterns that describe only the positive class and also discriminate it from the negative class. Thus we need only rules from the positive data $S^+$ and not from $S^-$. We use $\mathcal{A}_{S}$ to represent a complete set of patterns mined from $S^+$. $\mathcal{A}_S$ can be further divided into pattern pools indexed by lengths. A pool containing all patterns of length $l$ is denoted as $\mathcal{A}_S^{[l]}$. In this model, interpretability of a pattern  is determined by its length, so that the \textit{a priori} probability that a pattern of length $l$ is selected depends only on $l$.
We use a beta prior on the probability $p_l$ for the inclusion of a pattern in $\mathcal{A}_S^{[l]}$ to be in the pattern set $A$:
\small
\begin{equation}
 p_l \sim\text{Beta}(\alpha_l,\beta_l).
\end{equation}
The parameters $\{\alpha_l,\beta_l>0|l \in \{1,...,L\}\}$ on the priors control the expected number of patterns of each length in the pattern set.  Specifically, let $ A^{[l]}$ denote the set of patterns selected from $\mathcal{A}_S^{[l]}$, then the pattern set is represented by $A =
\cup_{l\in\{1,...\}}A^{[l]}$. Define $M_l = |A^{[l]}|$, 
we then have $E[M_l] = |A^{[l]}|\frac{\alpha_l}{\alpha_l + \beta_l}$. Therefore, if we favor short patterns, we could simply increase $\frac{\alpha_l}{\alpha_l + \beta_l}$ for smaller $l$ and decrease the ratio for larger $l$.

The pattern set $A^{[l]}$ is a collection of $M_l$ patterns independently selected from $\mathcal{A}_S^{[l]}$, $A^{[l]} \subseteq \mathcal{A}_S^{[l]}$. We integrate out the probability $p_l$ to get the probability of $A^{[l]}$:
\begin{align}
P(A^{[l]};\alpha_l,\beta_l) &= \int_{p_l} p_l^{M_l}(1-p_l)^{|\mathcal{A}_S^{[l]}|-M_l} \text{Beta}(p_l;\alpha_l,\beta_l)d(p_l) \notag \\
&=
\frac{\Gamma(\alpha_l + \beta_l)}{\Gamma(\alpha_l)\Gamma(\beta_l)}\frac{\Gamma(M_l+\alpha_l)\Gamma(|\mathcal{A}_S^{[l]}|-M_l+\beta_l)}{\Gamma(|\mathcal{A}_S^{[l]}|+\alpha_l+\beta_l)},
\end{align}
where the first line follows because each pattern is selected independently and the second line follows from integrating over the beta prior on $p_l$. Thus the probability of $A$ is:
\small
\begin{equation}
P(A;\theta_{\text{prior}}) = \prod_l^L P(A^{[l]};\alpha_l,\beta_l),
\end{equation}
where $\theta_{\text{prior}}=\{\alpha_l,\beta_l\}_l$. We usually choose $\alpha_l \ll \beta_l$ so that BOA tends to choose a smaller $M_l$ for each $l$.

\subsubsection{Poisson prior}
We introduce a different prior for the BOA model. Let $M$ denote the total number of patterns in $A$ and $L_m$ denote the lengths of patterns for $m\in\{1,...M\}$. There can be at least 0 and at most $|\mathcal{A}_S|$ patterns in a set, and for each pattern, the length needs to be at least 1 and at most the number of all attributes $J$. 
We first draw $M$ and $L_m$ from truncated Poisson distributions to decide the ``shape'' of a pattern set, then we fill in the patterns with literals. To generate a literal, we first  randomly select the attributes then randomly select values corresponding to each attribute. We use $v_{m,k}$ to represent the attribute index for the $k$-th literal in the $m$-th pattern, $v_{m,k} \in \{1,...J\}$. $K_{v_{m,k}}$ is the total number of values for attribute $v_{m,k}$. The generative process is as follows:
\begin{algorithmic}[1]
 \State Draw the number of patterns: $M \sim \text{Truncated-Poisson}(\lambda_M),M\in\{0,...|\mathcal{A}_S|\}$
 \For {$m \in\{1,...M\}$}
 \State Draw the number of conditions in $m$-th pattern: $L_m \sim  \text{Truncated-Poisson}(\lambda_L),L_m\in\{1,...J\} $
 \State Randomly select $L_m$ attributes from $J$ attributes without replacement
 \For {$k \in \{1,...L_m\}$}
 \State Uniformly at random select a value from $K_{v_{m,k}}$ values corresponding to attribute $v_{m,k}$
 \EndFor
 \EndFor
\end{algorithmic}

We define $\theta_{\text{prior}} = \{\lambda_M,\lambda_L \}$, and normalization constant $\omega(\lambda_M,\lambda_L)$, thus the probability of generating a pattern set $A$ is
\small
\begin{equation}
P(A;\theta_{\text{prior}}) = \omega(\lambda_M,\lambda_L)\text{Poisson}(M;\lambda_M) \prod_m^M \text{Poisson}(L_m;\lambda_L)\frac{1}{\binom{J}{L_m}}\prod_k^{L_m}\frac{1}{K_{v_{m,k}}}.
\end{equation}
\subsection{Likelihood}
Let $(\mathbf{x}_n,y_n)$ denote the $n$-th observation, let $f_A(\mathbf{x}_n)$ denote the classification outcome for $\mathbf{x}_n$, and let $y_n$ denote the observed outcome. Recall that $f_A(\mathbf{x}_n)=1$ if $X$ obeys any of the patterns $a \in {A}$.  We introduce likelihood parameter $\rho_+$ to govern the probability that an observation is a real positive class case $y_n=1$ when it satisfies the pattern set, and $\rho_-$ as the probability that $y_n=0$ when it does not satisfy the set.

The likelihood of data $S = \{\mathbf{x}_n,y_n\}_n$ given a pattern set $A$ and  parameters $\rho_+, \rho_-$ is thus:
\small
\begin{equation}
\label{eqn:not_marginalized_ll}
P(S|A,\rho_+,\rho_-) = \prod_n \rho_+^{f_A(\mathbf{x}_n)y_n}  (1-\rho_+)^{f_A(\mathbf{x}_n)(1-y_n)}{\rho_-}^{\left[1-f_A(\mathbf{x}_n)\right](1-y_n)}{(1-\rho_-)}^{\left[1-f_A(\mathbf{x}_n)\right]y_n},
\end{equation}
where the four components in formula (\ref{eqn:not_marginalized_ll}) represent four classification outcomes: true positive, false positive, true negative, and false negative.
We place beta priors over $\rho_+$ and $\rho_-$:
\small
\begin{equation*}
\rho_+ \sim \text{Beta}(\alpha_+,\beta_+), \quad\quad \rho_- \sim \text{Beta}(\alpha_-,\beta_-).\nonumber
\end{equation*}
Here, $\alpha_+,\beta_+,\alpha_-,\beta_-$ should be chosen such that
$E[\rho_+]$ and $E[\rho_-]$ are close to 1  which means the classification outcomes agree with the observed outcomes.
Integrating out 
$\rho_+$ and $\rho_-$ from the likelihood in
(\ref{eqn:not_marginalized_ll}), we get
\small
\begin{align}
P(S|A,\theta_{\text{likelihood}}) &= \int_0^1 P(S|A,\rho_+,\rho_-)\text{Beta}(\alpha_+,\beta_+)\text{Beta}(\alpha_-,\beta_-) d\rho_+ d\rho_- \notag \\
& = \frac{\Gamma(\alpha_+ + \beta_+)}{\Gamma(\alpha_+)\Gamma(\beta_+)}\frac{\Gamma\left(\sum_n f_A(\mathbf{x}_n)y_n+\alpha_+\right)\Gamma\left(\sum_nf_A(\mathbf{x}_n)(1-y_n)+\beta_+\right)}{\Gamma\left(\sum_nf_A(\mathbf{x}_n)+\alpha_++\beta_+\right)}\cdot  \notag \\ 
\frac{\Gamma\left(\alpha_- + \beta_-\right)}{\Gamma\left(\alpha_-\right)\Gamma(\beta_-)}&\frac{\Gamma\left(\sum_n \left(1-f_A(\mathbf{x}_n)\right)y_n+\beta_-\right)\Gamma\left(\sum_n \left(1-f_A(\mathbf{x}_n)\right)(1-y_n)+\alpha_-\right)}{\Gamma\left(\sum_n\left(1-f_A(\mathbf{x}_n)\right)+\alpha_-+\beta_-\right)},
\end{align} 
where $\theta_{\text{likelihood}} = \{\alpha_+,\beta_+,\alpha_-,\beta_-\}$.
The training data can be divided into four cases: true positives (TP = $\sum_nf_A(\mathbf{x}_n)y_n$), false positives (FP = $\sum_nf_A(\mathbf{x}_n)(1-y_n)$), true negatives (TN = $\sum_n\left(1-f_A(\mathbf{x}_n)\right)(1-y_n)$) and false negatives (FN = $\sum_n\left(1-f_A(\mathbf{x}_n)\right)y_n$). The above likelihood can be rewritten as:
\small
\begin{equation}
P(S|A,\theta_{\textrm{likelihood}}) =\frac{\Gamma(\alpha_+ + \beta_+)}{\Gamma(\alpha_+)\Gamma(\beta_+)}\frac{\Gamma(\text{TP}+\alpha_+)\Gamma(\text{FP}+\beta_+)}{\Gamma(\text{TP}+\text{FP}+\alpha_++\beta_+)}\cdot \frac{\Gamma(\alpha_- + \beta_-)}{\Gamma(\alpha_-)\Gamma(\beta_-)}\frac{\Gamma(\text{TN}+\alpha_-)\Gamma(\text{FN}+\beta_-)}{\Gamma(\text{TN}+\text{FN}+\alpha_-+\beta_-)}.
\end{equation}
We want to maximize the posterior, which is equivalent to maximizing the joint probability of $S$ and $A$.
\small
\begin{equation}
P(S,A;\theta) = P(S|A;\theta_{\textrm{likelihood}})P(A;\theta_{\textrm{prior}}),
\end{equation}
where $\theta=\{\theta_{\text{prior}},\theta_{\text{likelihood}}\}$ and $\theta_{\text{prior}}$ depends on which prior model is used.
\section{Approximate MAP Inference}
\label{sec:inference}
In this section, we describe a procedure for approximately solving for the maximum \emph{a posteriori} or MAP solution to the BOA model. Inference in the BOA model is challenging because finding the best model involves a search over exponentially many possible sets of patterns: since each pattern is a conjunction of literals, the number of patterns increases exponentially with the number of literals, and the number of sets of patterns increases exponentially with the number of patterns. We propose two variations of stochastic local search algorithms, with different notions of a ``neighboring'' solution. The first method searches the space by adding or removing a pattern at every iteration, and uses only pre-mined patterns. The second method searches the space by adding or removing a literal at every iteration. Both methods use a simulated annealing approach with moves designed to quickly explore promising solutions.
\vspace{-3mm}
\subsection{Simulated Annealing}
\label{sec:SA}
Maximizing the posterior is equivalent to minimizing the negative log joint probability:
\begin{equation}\label{eqn:objective}
E_S(A;\theta) = - \log P(S,A;\theta).
\end{equation}
We define $\Lambda_{S}$ as a complete set of all possible pattern sets:
\begin{equation}\label{eqn:lambda}
\Lambda_{S}:=\{A|A=\{a_1,a_2,...a_{|A|}\} \text{ where } a_i\in \mathcal{A}_S\}.
\end{equation}
Exhaustive evaluation of $A$ over all $\Lambda_S$ will not be feasible; if we were to brute force search for the best classifier out of the whole set of possible classifiers, this would involve evaluating all possible subsets of patterns on the training data, and for candidate pattern set $\mathcal{A}_S$, there are $|\Lambda_{S}| = 2^{|\mathcal{A}_S|}$ such subsets. Simulated annealing \cite{kirkpatrick1983optimization} presents an alternative, and is naturally suited to approximate optimization here. Our simulated annealing steps are similar to the Gibbs sampling steps used by \cite{LethamRuMcMa15,WangRu15} for rule-list models.

The search starts by randomly generating a pattern set. Then at each iteration, an example is randomly selected from the misclassified data points. If the example is positive, it means the current pattern set fails to cover it and we then find a neighboring solution that covers more data than the current solution, so we call the action ``COVERMORE''.  If the example is negative, it means the current pattern set covers the wrong data so we need to find a neighbor pattern set that covers less, and we call the action ``COVERLESS''. $A^{t+1}$ is generated from the current $A^t$ using one of the two actions. How the action is carried out on $A^t$ depends on the level where a ``neighbor'' is defined. In a \emph{pattern level} search, a change is made by either adding or removing a pattern; and in a \emph{literal level} search, a change is made by either adding or removing a literal. We will elaborate on the actions on the two levels later in this section. 
To help avoid local minima, we use a scoring function to evaluate all the neighboring solutions, select the best solution with probability $1-p$, and select a random solution with probability $p$. Since the objective is to minimize $E_S$, $E_S$ naturally becomes the scoring function to evaluate the neighboring solutions. To summarize:
\begin{itemize}
\compresslist
\item With probability $p$, move to a randomly selected neighboring position,
\item Otherwise, move from to neighboring position with minimum $E_S(A^{t+1};\theta)$.
\end{itemize}
Then the propoal $A^{t+1}$ is accepted with probability $\min\left\{1,\exp\left(-\frac{E_S(A^{t+1};\theta) - E_S(A^t;\theta)}{T(t)}\right)\right\},$ where $T(t)$ is the temperature and it follows a cooling schedule $T(t) = \frac{T_0}{\log(1+t)}$.

We repeat the search three times, from three random starting points, and we select the solution with the highest MAP. We present the general search in Algorithm 1, where the user can choose to do a pattern level search (see \ref{sec:patternlevel}) or a literal level search (see \ref{sec:literallevel}).

\begin{algorithm}[h!]
\caption{Simulated Annealing for BOA}
\begin{algorithmic}[1]
\Procedure{SLSearch}{\textit{maxSteps}, $p$, $level$}
\State $A^{t} \leftarrow$ a randomly generated pattern set,
\State \textit{step} $\leftarrow 0$, 
\State $S_\epsilon \leftarrow$ examples misclassified by $A^t$
 \While{\textit{step}<\textit{maxSteps} and $S_\epsilon\neq \emptyset$}
     \State $step \leftarrow step + 1$
     \State $S_\epsilon \leftarrow$ misclassified examples by $A^t$
     \State $ex \leftarrow$ a random example drawn from $S_\epsilon$
     \If{$ex$ is a positive example}
\State $A^{t+1} \leftarrow \text{COVERMORE}(level, A^t, p)$
        \Else
\State $A^{t+1} \leftarrow \text{COVERLESS}(level, A^t, p)$
        \EndIf
\State $A_\text{max} \leftarrow \max(A_\text{max},A^{t+1})$
\State $\alpha = \min\left\{1,\exp\left(-\frac{E_S(A^{t+1};\theta) - E_S(A^t;\theta)}{T(t)}\right)\right\}$
\State  \begin{equation*}
A^{t+1}  \leftarrow   \begin{cases}
        A^{t+1}, \text{  with probability }\alpha \\
        A^{t}, \text{  with probability }1-\alpha \\
        \end{cases}
\end{equation*}
\EndWhile
\State return $A_{\text{max}}$
\EndProcedure
\end{algorithmic}
\end{algorithm}

\subsection{Pattern Level Stochastic Search}\label{sec:patternlevel}
For stochastic search over patterns, neighboring solutions are ``one-pattern-different'' from the current set. The simulation chain moves to these positions by adding or removing a pattern from the current one. Therefore the two actions are the following:
\begin{itemize}
\compresslist
\item COVERMORE(``pattern'', $A_t$, $p$) 
\begin{itemize}
\item With probability $p$, add a random pattern to $A^t$.
\item Else, evaluate the objective $E_S$ for all neighboring solutions where a pattern is added to $A^t$ and choose the one with the minimum score.
\end{itemize}
\item COVERLESS(``pattern'', $A_t$, $p$) 
\begin{itemize}
\item With probability $p$, remove a random pattern from $A^t$.
\item Else, evaluate the objective $E_S$ for all neighboring solutions where a pattern is removed from $A^t$ and choose the one with the minimum score.
\end{itemize}
\end{itemize}

We note that any reasonably accurate sparse classifier should contain largely accurate patterns.  Rather than considering all patterns (exponential in the number of attributes), we use only the pre-mined patterns.  
To efficiently search for the MAP solution, we require a minimum support to limit the number of patterns that are generated. This will greatly reduce the computational complexity and we will show in Section \ref{sec:bounds} that filtering out these patterns does not affect the MAP joint probability. 



We consider both positive associations (e.g., $x_j$=`blue') and negative associations ($x_j$=`not green') as literals. (The importance of negative literals is stressed, for instance, by \cite{brin1997beyond,wu2002mining, teng2002mining}.) We then mine for frequent patterns within the set of positive observations
$S^+$. To do this, we use the FP-growth algorithm \cite{borgelt2005implementation}, 
which can in practice be replaced with any desired frequent pattern-mining method.
Even when we restrict the length of patterns and the minimum support, the number of generated patterns could still be too large to handle. (For example, a million patterns are generated for one of the advertisement datasets the minimum support is 5\% and the maximum length is 3). Therefore, we wish to use a second criterion to screen for the most potentially useful $M_0$ patterns. We first filter out patterns on the lower right plane of ROC space, i.e., their false positive rate is greater than true positive rate. Then we use \textit{information gain} to screen patterns, similarly to other works \cite{chen2006new,cheng2007discriminative}. For a pattern $a$, the information gain is $\text{InfoGain}(S|a) = H(S) - H(S,a ),$
where $H(S)$ is the entropy of the data and $H(S, a)$ is the entropy of data that split on pattern $a$. Given a dataset $S$, entropy $H(S)$ is constant; therefore our screening technique chooses the $M_0$ patterns that have the smallest $H(S,a)$, where $M_0$ is user-defined. 
\begin{figure}[h]
\centering
\includegraphics[width=0.45\textwidth]{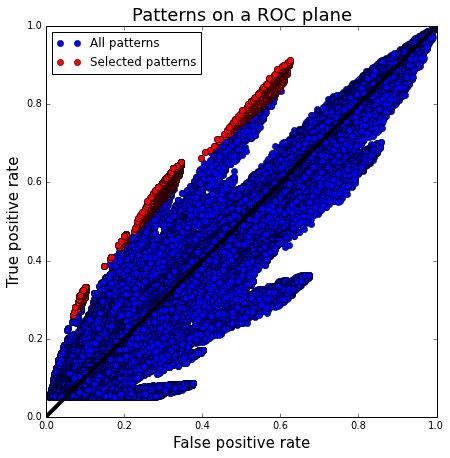}
\caption{All patterns and selected patterns on a ROC plane}
\label{fig:roc_patterns}
\end{figure}
We illustrate the effect of screening on one of our advertisement data sets. We mined all patterns with minimum support 5\% and maximum length 3. For each pattern, we computed its true positive rate and false positive rate on the training data, and plotted it as a dot in Figure \ref{fig:roc_patterns}. The top $5000$ patterns with highest information gains are colored in red, and the rest are in blue. As shown in the figure,  information gain indeed selected good patterns as they are closer to the upper left corner in ROC space. For many applications, this screening technique is not needed, and we can simply use the entire set of pre-mined rules.

\subsection{Literal Level Stochastic Search}\label{sec:literallevel}
For search on literal level, a neighboring solution is generated by either adding or removing a literal from the current pattern set. If the example is a positive case, it means the current pattern set fails to cover it. In order to cover this example, we either remove a literal from a pattern or add a literal as a new pattern. Both options increase the support of the pattern set to cover more data points. If the example is a negative case, it means the current pattern set covers the wrong data so we need to either remove an existing pattern or add a condition to a pattern to make it not cover the example. 

\begin{itemize}
\compresslist
\item COVERMORE(``literal'', $A_t$, $p$)
\begin{itemize}
\item With probability 0.5, do as follows: with probability $p$, remove a random literal from a random pattern; else, evaluate the objective $E_S$ for all neighboring solutions where a literal is removed and choose the one with the minimum score.
\item With probability 0.5, do as follows: with probability $p$, add a random literal as a new pattern; else, evaluate the objective $E_S$ for all neighboring solutions where a literal is added as a new pattern, choose the one with the minimum score.
\end{itemize}
\item COVERLESS(``literal'', $A_t$, $p$) 
\begin{itemize}
\item With probability 0.5, do as follows: with probability $p$, add a random literal to a random pattern; else, evaluate the objective $E_S$ for all neighboring solutions where a literal is added to a pattern in the set and choose the one with the minimum score.
\item  With probability 0.5, do as follows: with probability $p$, remove a random pattern, else, evaluate the objective $E_S$ for all neighboring solutions where a pattern is removed from the current set and choose the one with the minimum score.
\end{itemize}  
\end{itemize}

The second action in COVERLESS removes a pattern. Intuitively it might make more sense to remove one literal at a time rather than removing a pattern, but actually, removing a whole pattern is often a more gradual change to the model than removing a single literal. A large pattern (consisting of several constraints on the input variables) affects fewer points than a small pattern (consisting of a single constraint). Thus, removing a whole pattern can often be a more ``local" move than removing a single literal.


This algorithm has some advantages over pattern level stochastic search, in that it does not need to pre-define a maximum length of patterns and does not need to generate patterns beforehand. This allows us to search the possibly huge space of patterns without enumerating all of them.

\section{Guarantees on the MAP BOA Models from the Priors} \label{sec:bounds}
We will show that the inclusion of the prior (which was designed to help with interpretability), provably assists with both computation and generalization performance. 
Recall the objective defined in Equation (\ref{eqn:objective}): $E_S(A;\theta) = - \log P(S,A;\theta).$ The goal is to find an MAP pattern set $A^*$ that minimizes $E_S$, which is equivalent to finding a MAP solution. 
We show that depending on the prior parameters and on the size of the data, we have deterministic (i) upper bounds on the sizes of MAP BOA models, and (ii) lower bounds on the support of rules we need to construct MAP BOA models. This means that practically, we need only mine and optimize over rules of a certain support level to obtain a MAP solution, which exponentially reduce the size of the BOA model computation. These bounds also directly lead to better generalization bounds on predictive performance. The proofs of all theorems in this section are in Appendix \ref{SectionProofs}.


We first provide an upper bound for the size of BOA models that depends on the priors. That is, when the priors are chosen to favor smaller models, we can place an explicit guarantee on the size of the MAP BOA model. This bound depends explicitly on the prior parameters, the number of observations and the number of attributes. The reason that the size of a MAP solution is bounded is that the prior places a penalty large enough so that the likelihood cannot overwhelm it.
We first study the BOA-Poisson model. 

\begin{thm}\label{Theorem1}
Take a BOA-Poisson model with parameters $\theta =\left \{\alpha_+,\beta_+\alpha_-,\beta_-,\lambda_M,\lambda_L\right\}$, where $\alpha_+$, $\beta_+$, $\alpha_-$,$\beta_-$, $\lambda_M,\lambda_L \in \mathbb{N}^+$. The dataset $S$ has $J$ attributes, where the $j$-th attribute has $K_j$ values for $j \in \{1,...,J\}$. Define $A^* \in \arg\min_A E_S(A)$ and $M^* := |A^*|$. If $\frac{e^{-\lambda_L}\left(\frac{\lambda_L}{2}\right)^J}{\Gamma(J+1)}\leq 1$, Then 
\small
\begin{equation*}
M^* \leq \lambda_M + \frac{\log\left( \frac{\Gamma(\alpha_-+\beta_-)}{\Gamma(\alpha_-)\Gamma(\beta_-)}\frac{\Gamma(|S^-|+\alpha_-)\Gamma(|S^+|+\beta_-)}{\Gamma(|S|+\alpha_-+\beta_-)}\cdot\frac{\lambda_M!}{\left( \frac{e^{-\lambda_L}\lambda_M \max \big\{\left(\frac{\lambda_L}{2}\right)\Gamma(J),\left(\frac{\lambda_L}{2}\right)^J \big\}}{\Gamma(J+1)}\right)^{\lambda_M}}\right)}{\log \left(\frac{e^{-\lambda_L}\lambda_M \max \big\{\left(\frac{\lambda_L}{2}\right)\Gamma(J),\left(\frac{\lambda_L}{2}\right)^J \big\}}{\Gamma(J+1)(\lambda_M+1)}\right)}.
\end{equation*}
\end{thm}

As we show in the proof, $ \frac{\Gamma(\alpha_-+\beta_-)}{\Gamma(\alpha_-)\Gamma(\beta_-)}\frac{\Gamma(|S^-|+\alpha_-)\Gamma(|S^+|+\beta_-)}{\Gamma(|S|+\alpha_-+\beta_-)}$ is the likelihood of data given an empty set, i.e., $P(S|\emptyset;\theta)$. As $P(S|\emptyset;\theta)$ increases, the bound becomes smaller, which means the model's maximum possible size is smaller. Intuitively, if an empty set already achieves high likelihood, adding rules will often hurt the prior term and achieve little gain on the likelihood. Assuming the ratio of $S^+$ and $S^-$ stays the same, as the amount of data increases, the term $\frac{\Gamma(|S^-|+\alpha_-)\Gamma(|S^+|+\beta_-)}{\Gamma(|S|+\alpha_-+\beta_-)}$ becomes smaller and $\log \frac{\Gamma(|S^-|+\alpha_-)\Gamma(|S^+|+\beta_-)}{\Gamma(|S|+\alpha_-+\beta_-)}$ becomes more negative. Since the denominator is negative and other terms stay the same, the bound becomes larger. The data overwhelm the prior on the size of the model.

We have a similar upper bound for the size of a MAP BOA-BetaBinomial pattern set.
\begin{thm}\label{Theorem2}
Take a BOA-BetaBinomial model with parameters $$\theta = \left\{L,\alpha_+,\beta_+,\alpha_-,\beta_-,\lambda_M,\lambda_L,\{\mathcal{A}_S^{[l]},\alpha_l,\beta_l\}_{l=1,..L}\right\},$$ where $L,\lambda_M,\lambda_L,\alpha_+,\beta_+,\alpha_-,\beta_-,\{\alpha_l,\beta_l\}_{l=1,...L} \in \mathbb{N}^+$. Define $A^* \in \arg\min_A E_S(A)$, and $M^*:=|A^*|$. Whenever $\alpha_l<\beta_l$, we have: 
\begin{equation*}
M^* \leq \sum_l^L\frac{\log  \left(\frac{\Gamma(\alpha_-+\beta_-)}{\Gamma(\alpha_-)\Gamma(\beta_-)}\frac{\Gamma(|S^-|+\alpha_-)\Gamma(|S^+|+\beta_-)}{\Gamma(|S|+\alpha_-+\beta_-)}\right)}{\log \frac{|\mathcal{A}_S^{[l]}|+\alpha_l-1}{|\mathcal{A}_S^{[l]}|+\beta_l-1}}.
\end{equation*}
\end{thm}
Similar intuition holds as for the BOA-Poisson model: the dependence on the number of observations is the same for the two bounds. Additionally, when $\alpha_l$ is set to be small and $\beta_l$ is set to be large, the bound is smaller so we are guaranteed to choose a smaller number of rules overall. This is consistent with users' expectation when they set $\frac{\alpha_l}{\alpha_l+\beta_l}$ to be small, as explained in Section~\ref{sec:betabinomial}. 

%

In the BOA-BetaBinomial model, we pre-mine patterns to filter out those with small support. Let us not use information gain, and discuss the implications of using only low support rules. This is equivalent to the statistical assumption that the set of pre-mined patterns are sufficient to produce with posterior approximately the same as the MAP model. We will show something stronger than this. Namely, we show that finding a globally MAP pattern list (among all possible pattern lists) is equivalent, under weak conditions, to the pattern list found from only the pre-mined patterns. This yields a major computational benefit, as it tells us that we need only mine and optimize over rules of a certain minimum support in order to find the MAP solution. This eliminates an enormous number of rules and decreases the search space substantially. The stronger the prior, the more tractable the computation for this model.

We define a subset of $\mathcal{A}_S$ containing patterns of support at least $C$ as:
\small
\begin{equation*}
\mathcal{A}_{S}^C:=\{a \in \mathcal{A}_S:\text{supp}_{S}(a)\geq C\}.
\end{equation*}
$\mathcal{A}_{S}^C$ is fully enumerated in the rule mining step.
Ideally we would like to optimize over elements of $\mathcal{A}_S$, but it is only practical to optimize over $\mathcal{A}_{S}^C$. We will thus prove that it is sufficient to do this in order to find a MAP solution; the MAP solution has only patterns that come from $\mathcal{A}_{S}^C$.
 We define $\Lambda_{S}^C$ as the set of all possible BOA models with support at least $C$ for each pattern:
$$\Lambda_{S}^C:=\{A|A=\{a_1,a_2,...a_{|A|}\} \text{ where } a_i\in \mathcal{A}_S^C\}.$$
If $C=0$, $\Lambda_{S}^0$ is the set of all possible BOA models, $\Lambda_{S}$ as defined in (\ref{eqn:lambda}). 


\begin{thm}\label{Theorem3}
Take a BOA-BetaBinomial model with parameters $$\theta = \left\{L,\alpha_+,\beta_+,\alpha_-,\beta_-,\lambda_M,\lambda_L,\{\mathcal{A}_S^{[l]},\alpha_l,\beta_l\}_{l=1,..L}\right\},$$ where $L,\lambda_M,\lambda_L,\alpha_+, \beta_+,\alpha_-,\beta_-,\alpha_l,\beta_l \in \mathbb{N}^+$. When $\frac{|S^+|+\alpha_++\beta_+-1}{|S^+|+\alpha_+-1}\frac{\beta_-}{|S^-|+\alpha_-+\beta_-}\leq1$, $\alpha_l<\beta_l$ for $l = 1,2...L$, and $$C  \leq \frac{\log\underset{l}{\min}\left(\frac{|\mathcal{A}_S^{[l]}|-m_{l}+\beta_{l}}{m_{l}-1+\alpha_{l}}\right)}{\log \frac{|S^+|+\alpha_+-1}{|S^+|+\alpha_++\beta_+-1}\frac{|S^-|+\alpha_-+\beta_-}{\beta_-}},$$ where $m_l$ is the upper bound computed in Theorem \ref{Theorem2}, then
$$\underset{A\in\Lambda_S}{\arg\min}E_S(A) \subseteq \underset{A\in\Lambda_{S}^{C}}{\arg\min}E_S(A).$$
\end{thm}
This theorem states that the quantity $\underset{A\in\Lambda_S}{\arg\min}E_S(A)$ on the left (which is computationally impossible to compute in practice) has the same set of minimizers as the quantity $\underset{A\in\Lambda_{S}^{C}}{\arg\min}E_S(A)$ on the right (which is what we would compute in practice). It states that we need only to mine patterns of support of at least C. If we solve for a minimizer of $E_S$ on these mined patterns, it is a global minimizer of $E_S$ across all pattern sets. This provides strong computational motivation for pre-mining patterns, since there are now weak conditions under which the ``approximation" of using pre-mined patterns is not an approximation at all.
We have a similar bound on the support for BOA-Poisson models.
\begin{thm}\label{Theorem4}
Take a BOA-Poisson model with parameters $\theta =\left \{\alpha_+,\beta_+\alpha_-,\beta_-,\lambda_M,\lambda_L\right\}$, where $\alpha_+$, $\beta_+$, $\alpha_-$,$\beta_-$, $\lambda_M,\lambda_L \in \mathbb{N}^+$. When $\frac{|S^+|+\alpha_++\beta_+-1}{|S^+|+\alpha_+-1}\frac{\beta_-}{|S^-|+\alpha_-+\beta_-}\leq1$, and $$C  \leq \frac{\log \left(\frac{\Gamma(J+1)}{\lambda_M e^{-\lambda_L}\max \left\{\left(\frac{\lambda}{2}\right)\Gamma(J),\left(\frac{\lambda}{2}\right)^J \right\}} \right)}{\log \frac{|S^+|+\alpha_+-1}{|S^+|+\alpha_++\beta_+-1}\frac{|S^-|+\alpha_-+\beta_-}{\beta_-}},$$ then
$$\underset{A\in\Lambda_S}{\arg\min}E_S(A) \subseteq \underset{A\in\Lambda_{S}^{C}}{\arg\min}E_S(A).$$
\end{thm}

So far our model has been focusing on properties of the \emph{a posteriori} model. We now turn to generalization performance. The following result is algorithm independent. The true risk of a pattern set $A$ is defined as:
\small
\begin{equation*}
R^{\text{true}}(A) = \mathbb{E}_{(X,Y)\sim D}(f_{A}(X)\neq Y), 
\end{equation*}
where classifier $f_A(X)$ equals one when $X$ obeys one of the patterns in $A$, and $f_A(X)$ equals zero otherwise. The true risk is the standard expected misclassification error.
\begin{thm}\label{Theorem5}
Consider $\Lambda^\mathcal{F}$ which is the set of BOA models parameterized by $\{\rho_+,\rho_-\}$, $\rho_+$, $\rho_-$ $< 1/2$, where the number of patterns of $A \in\Lambda^\mathcal{F}$ obeys $|A|\leq M_\text{upper}$. With probability $1-\delta$, for all $A \in \Lambda^\mathcal{F}$,
\begin{equation*}
R^{\text{true}}(A) \leq   \frac{\log P(S|A;\rho_+,\rho_-)}{N\log\frac{1}{2}}  + \sqrt{\frac{\sum_{m=1}^{M_{\text{upper}}}{\prod_j^{J}(K_j +1) \choose m}+\log \frac{1}{\delta}}{2N}}.
\end{equation*}
\end{thm}
This theorem states that the true risk is upper bounded by the empirical risk of data $S$, in particular the likelihood, and a complexity term. The complexity term increases with $\prod_j^{J}(K_j +1)$, which is the number of all patterns. 
The value of $M_{\text{upper}}$ can come from either Theorem \ref{Theorem1} or Theorem \ref{Theorem2}.
\section{Simulation Studies}
In this section, we present simulation studies to show that if data are generated from a fixed pattern set, our simulated annealing procedure can recover it with high probability. We also provide convergence analysis on simulated data sets to show that our model can achieve the MAP solution in a relatively short time.

\subsection{Performance variation with different parameters}
Given observations $\{\mathbf{x}_n\}_{n=1,...N}$ and assume there exists a true pattern set comprised of $m$ patterns that classifies $x_n$, and generates the outcome $y_n$. We want to show that simulated annealing can discover this pattern set efficiently. Let there be a collection of patterns $\{a_j\}_{j=1,...M}$ mined from the observations, and we can construct a binary matrix where the entry on $n$-th row and $j$-th column is the Boolean function $h(x_n,a_j)$, which represents if the $n$-th observation satisfies the $j$-th patterns. We need only to simulate this binary matrix to represent the observations without losing generality. Each entry is set to 1 independently with probability 0.1. Here are the most important variables in this simulation study:
\begin{itemize}
\compresslist
\item $M$: the number of candidate patterns  
\item $m$: the number of patterns in a true pattern set
\item $N$: the number of observations in a data set
\end{itemize}
The binary matrix representing the data set has size $N\times M$. We assume all patterns have the same length so we can ignore the priors. 
In our experiments, a true pattern set was generated by randomly selecting $m$ patterns to form a pattern set. We used edit distance between the true pattern set and a generated pattern set as the performance measure. For each number of iterations from 5000, 10000 and 20000, we repeated experiments in the simulation 100 times. For each recovery problem, we then used simulated annealing as described in Section \ref{sec:SA} with three different starting points. We reported the mean performance in Table~\ref{tab:simulations}.
\paragraph{Performance with size of data set, $N$.}
In the first study, we set $m = 5$, $M = 1000$, 
and chose the sample size $N$ from $\{100,500,1000,2000,3000,4000\}$. 
The edit distance was computed for each of the 100 replicates and the means are reported in Table~\ref{tab:simulations}. Our results show that as the number of iterations increases, the true pattern sets were recovered with higher probability. However, the number of observations $N$ did not have a large influence on the result for $N$ approximately greater than 500. 
The accuracy at $N=500$ is similar to the accuracy at $N=4000$. This result is quite intuitive since simulated annealing searches over the pattern space, and likely finds the same solution once $N$ is sufficiently large. 
\begin{table}[h]
\centering
\begin{tabular}{ll|c|c|c}
\toprule                                                                                          &        & \multicolumn{1}{l|}{\#of iteration=5000} & \multicolumn{1}{l|}{\#of iteration=10000} & \multicolumn{1}{l}{\#of iteration=20000} \\ \hline
\multicolumn{1}{c}{\multirow{6}{*}{\begin{tabular}[c]{@{}c@{}}M=1000\\ m=5\end{tabular}}} & N=100  & 3.13                                     & 2.65                                      & 2.31                                      \\
\multicolumn{1}{c}{}                                                                      & N=500  & 1.26                                     & 0.51                                      & 0.05                                      \\
\multicolumn{1}{c}{}                                                                      & N=1000 & 1.43                                     & 0.62                                      & 0.05                                      \\
\multicolumn{1}{c}{}                                                                      & N=2000 & 1.37                                     & 0.47                                      & 0.06                                      \\
\multicolumn{1}{c}{}                                                                      & N=3000 & 1.37                                     & 0.41                                      & 0.07                                      \\
\multicolumn{1}{c}{}                                                                      & N=4000 & 1.43                                     & 0.61                                      & 0.04                                      \\ \hline
\multirow{4}{*}{\begin{tabular}[c]{@{}l@{}}N=2000\\  m=5\end{tabular}}                    & M=200  & 0.00                                     & 0.00                                      & 0.00                                      \\
                                                                                          & M=500  & 0.74                                     & 1.16                                      & 0.00                                      \\
                                                                                          & M=1000 & 1.37                                     & 0.47                                      & 0.06                                      \\
                                                                                          & M=2000 & 2.28                                     & 1.47                                      & 0.84                                      \\ \hline
\multirow{5}{*}{\begin{tabular}[c]{@{}l@{}}N=2000\\ M=1000\end{tabular}}                  & m=1    & 0.01                                     & 0.00                                      & 0.00                                      \\
                                                                                          & m=2    & 0.23                                     & 0.08                                      & 0.00                                      \\
                                                                                          & m=4    & 0.81                                     & 0.38                                      & 0.02                                      \\
                                                                                          & m=6    & 2.09                                     & 0.84                                      & 0.12                                      \\
                                                                                          & m=8    & 4.72                                     & 4.49                                      & 1.8                                      
\end{tabular}
\caption{Mean edit distances to true patten sets with different $N, M$ and $m$.}
\label{tab:simulations}
\end{table}
\paragraph{Performance with size of pattern space, $M$.}
In the second study, we set $m = 5, N = 2000$, 
and chose the pattern size $M$ from $\{100, 200, 500,1000,2000\}$. We repeated the above procedure and reported the mean over 100 replicates in Table~\ref{tab:simulations}. The number of possible pattern sets of patterns is $\mathcal{O}(2^M)$; therefore as $M$ increases, searching the space becomes difficult for simulated annealing. (This does not mean, however, that prediction performance will suffer; as we increase the number of iterations, the mean edit distance decreases.) We can compensate for larger $M$ by running the simulation for longer times in order to recover the underlying pattern.
\paragraph{Performance with size of true pattern set, $m$.}
In the third study, we set $N = 2000, M = 1000$ and chose the size of the true pattern $m$ within $\{1,2,4,6,8\}$. Table~\ref{tab:simulations} shows that as the number of patterns increases, it becomes harder for the model to recover the true pattern set; however, performance improves over simulated annealing iterations. 
%
\subsection{Runtime analysis}
\begin{figure}[h]
\centering
\includegraphics[width=0.8\textwidth]{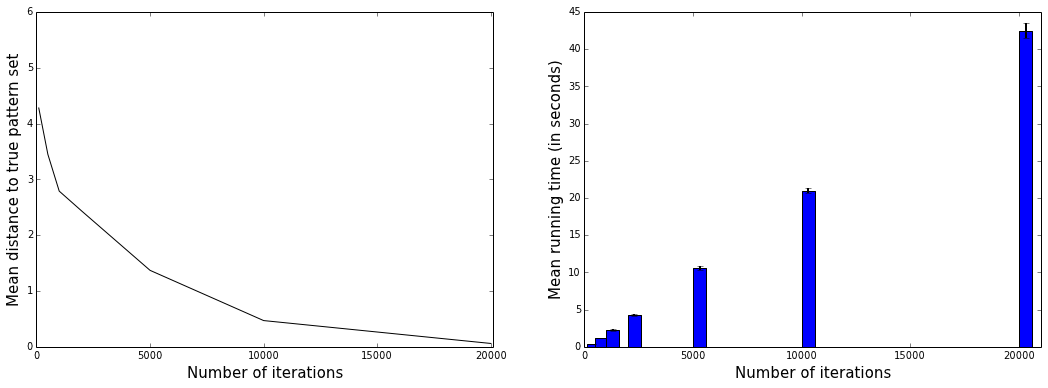}
\caption{Converence of mean edit distance and running time with number of iterations.}
\label{fig:convergence}
\end{figure}

We show how efficiently the performance improves as the our algorithm runs. We set the size of the data $N$ to be 2000 and the size of the true pattern set $m$ to be 5. We then ran simulated annealing and recorded the output at steps 100, 500, 1000, 2000, 5000, 10000 and 20000. We repeated this procedure 100 times and plotted the mean and variance of edit distances to true pattern sets in Figure~\ref{fig:convergence}, along with running times in seconds. The algorithm is also very fast. Running times were less than one minute, even for 20000 iterations. 

\section{Experiments}
We test our model on mobile advertisement datasets that we collected as well as other publicly available datasets. 
In situations where the ground truth consists of deterministic rules (similarly to the simulation study), our method tends to perform better than other popular machine learning techniques.
\subsection{Experiments on Mobile Advertisement Datasets}
For this experiment, our goal was to determine the feasibility of an in-vehicle recommender system that provide coupons for local businesses. 
The coupons would be targeted to the user in his/her particular context. 
Our data were collected on Amazon Mechanical Turk via a survey that we will describe shortly. 
We used Turkers with high ratings (95\% or above) 
and used two random questions with easy answers to reject surveys submitted by workers who were not paying attention. 
Out of 752 surveys, 652 were accepted, which generated 12684 data cases (after removing rows containing missing attributes).

The prediction problem is to predict if a customer is going to accept a coupon for a particular venue, considering demographic and contextual attributes. Answers that the user will drive there `right away' or `later before the coupon expires' are labeled as `Y = 1' and answers `no, I do not want the coupon' are labeled as `Y=0'. We are interested in investigating 5 types of coupons: bars, takeaway food restaurants, coffee houses, cheap restaurants (average expense below \$20 per person), expensive restaurants (average expense between \$20 to \$50 per person).  In the first part of the survey, we asked users to provide their demographic information and preferences, and in the second part, we described 20 different driving scenarios (see an example in the appendix) to each user along with additional context information and coupon information. We then asked the user if s/he will use the coupon. 

For categorical attributes, each attribute-value pair was directly coded into a literal. Using marital status as an example, `marital status is single' is converted into (MaritalStatus: Single), (MaritalStatus: Not Married partner), and (MaritalStatus: Not Unmarried partner), (MaritalStatus: Not Widowed). For discretized numerical attributes, the levels are ordered, such as: age is `20 to 25', or `26 to 30', etc; 
each attribute-value pair was converted into two literals, each using one side of the range. For example, age is `20 to 25' was converted into (Age:>=20) and (Age:<=25). Then each literal is a half-space defined by threshold values. See the appendix for a full description of attributes.

We will show that BOA does not lose too much accuracy on the mobile advertisement data sets (with respect to the highly complicated black box machine learning methods) even though we restricted the lengths of patterns and the number of patterns to yield very sparse models. We compared with other classification algorithms C4.5, CART, random forest, linear lasso, linear ridge, logistic lasso, logistic ridge, and SVM, which span the space of widely used methods that are known for interpretability and/or accuracy. The decision tree methods are representatives of the class of greedy and heuristic methods (e.g., \cite{ma1998integrating, li2001cmar, yin2003cpar,chen2006new,cheng2007discriminative,McCormickRuMa12,RudinLeMa13,MalioutovVa13, pollack1988pediatric, FriedmanFi99, gaines1995induction, cohen1995fast}) that yield interpretable models (though in many cases decision trees are often too large to be interpretable). For all experiments, we measured out-of-sample performance using AUC (the Area Under The ROC Curve) from 5-fold testing where the MAP BOA from the training data was used to predict on each test fold.\footnote{We do not perform hypothesis tests, as it is now known that they are not valid due to reuse of data over folds.}
We used the RWeka package in R for the implementations of the competing methods and tuned the hyperparameters using grid search in nested cross validation. 
For the pattern mining step for BOA-BetaBinomial models, we set the minimum support to be 5\% and set the maximum length of patterns to be 3. We used information gain to select the best 5000 patterns to use for BOA. We ran simulated annealing for 50000 iterations to obtain a pattern set. 
In the table, BOA1 represents BOA-BetaBinomial and BOA2 represents BOA-Poisson.

\begin{table*}[h]
\centering
\small
\begin{tabular}{lccccc}
\toprule
& Bar           & \begin{tabular}[c]{@{}c@{}}Takeaway\\ Food\end{tabular} & Coffee House  & \begin{tabular}[c]{@{}c@{}}Cheap\\ Restaurant\end{tabular} & \begin{tabular}[c]{@{}c@{}}Expensive \\ Restaurant\end{tabular} \\ \toprule
\textbf{BOA1}            & 0.744 (0.021) & 0.672 (0.005) & 0.753 (0.010) & 0.736 (0.022)                                                        & 0.705 (0.025)                                                   \\
\textbf{BOA2}           & 0.756 (0.009) & 0.637 (0.023) & 0.756 (0.007) & 0.736 (0.019)               
& 0.707 (0.030)                                                   \\
C4.5           & 0.757 (0.015) & 0.602 (0.051) & 0.751 (0.018) & 0.692 (0.033)                                                        & 0.639 (0.027)                                                   \\
CART           & 0.772 (0.019) & 0.615 (0.035) & 0.758 (0.013) & 0.732 (0.018)                                                        & 0.657 (0.010)                                                   \\
RF & 0.798 (0.016) & 0.640 (0.036) & 0.815 (0.010) & 0.700 (0.022)                                                        & 0.689 (0.010)                                                   \\
Lin-Lasso   & 0.795 (0.014) & 0.673 (0.042) & 0.786 (0.011) & 0.769 (0.024)                                                        & 0.706 (0.017)                                                   \\
Lin-Ridge   & 0.795 (0.018) & 0.671 (0.043) & 0.784 (0.012) & 0.769 (0.020)                                                        & 0.706 (0.020)                                                   \\
Logi-Lasso & 0.796 (0.014) & 0.673 (0.042) & 0.787 (0.011) & 0.767 (0.024)                                                        & 0.706 (0.016)                                                   \\
Logi-Ridge & 0.793 (0.018) & 0.670 (0.042) & 0.783 (0.011) & 0.768 (0.021)                                                        & 0.705 (0.020)                                                   \\
SVM            & 0.842 (0.018) & 0.735 (0.031) & 0.845 (0.007) & 0.799 (0.022)                                                        & 0.736 (0.022)       \\
\bottomrule                                           
\end{tabular}
\caption{AUC comparison for mobile advertisement data set, means and standard deviations over folds are reported.}
\label{tab:auc_coupon}
\end{table*}

We considered five separate coupon prediction problems, for different types of coupons. The AUC's for BOA and baseline methods for all five problems are reported in Table~\ref{tab:auc_coupon}. The BOA classifier, while restricted to produce sparse disjunctions of conjunctions, tends to perform almost as well as the black box machine learning methods, and outperforms the decision tree algorithms. The linear modeling methods were cross-validated, so even when they are not restricted to be sparse, the sparse BOA models perform comparably.

\subsection{Interpretability of results}
In practice, for this particular application, the benefits of interpretability far outweigh small improvements in accuracy. An interpretable model can be useful to a vender choosing whether to provide a coupon and what type of coupon to provide, it can be useful to users of the recommender system, and it can be useful to the designers of the recommender system to understand the population of users and correlations with successful use of the system.
As discussed earlier, \textit{or}'s of \textit{and}'s classifiers are particularly useful for representing consumer behavior. 

\begin{figure}
\centering
\hfill
\subfigure[Coupons for bars]{\includegraphics[width=0.42\textwidth]{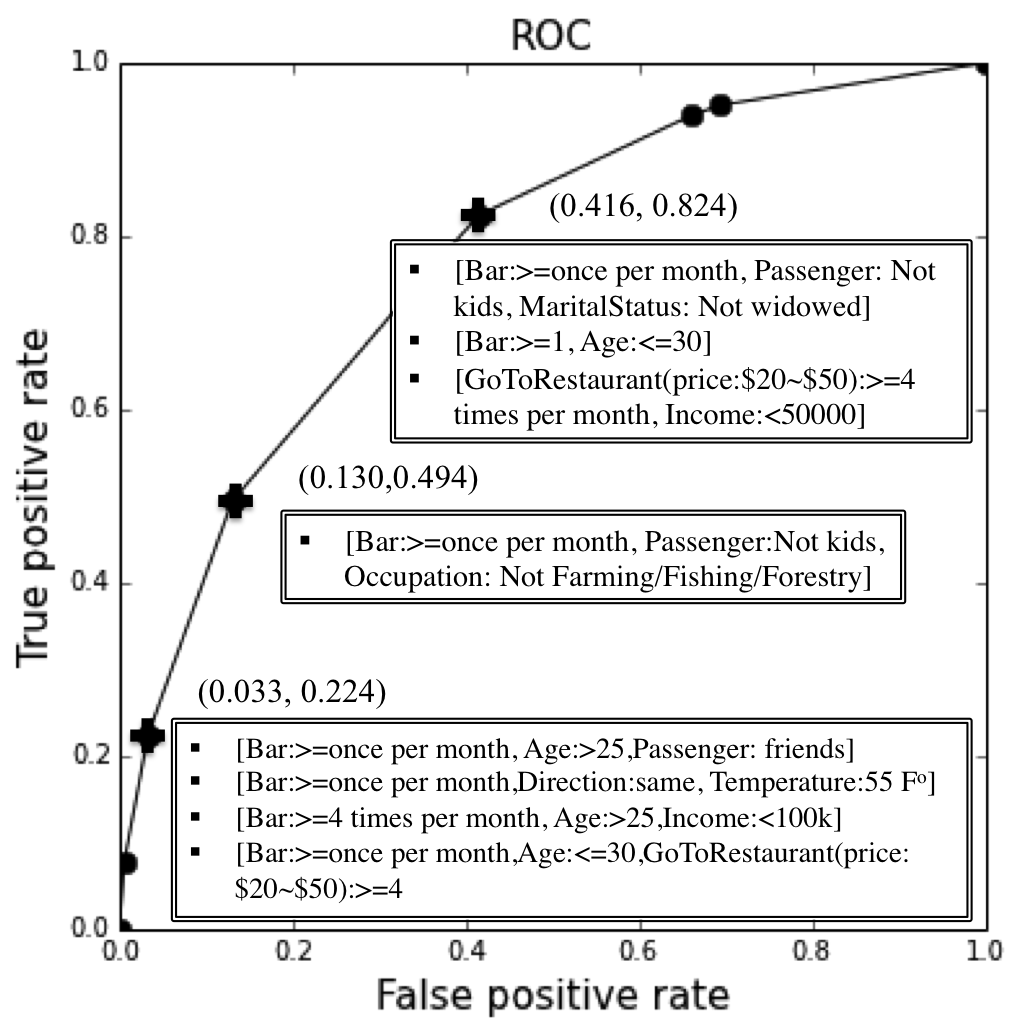}}
\hfill
\subfigure[Coupons for coffee houses]{\includegraphics[width=0.42\textwidth]{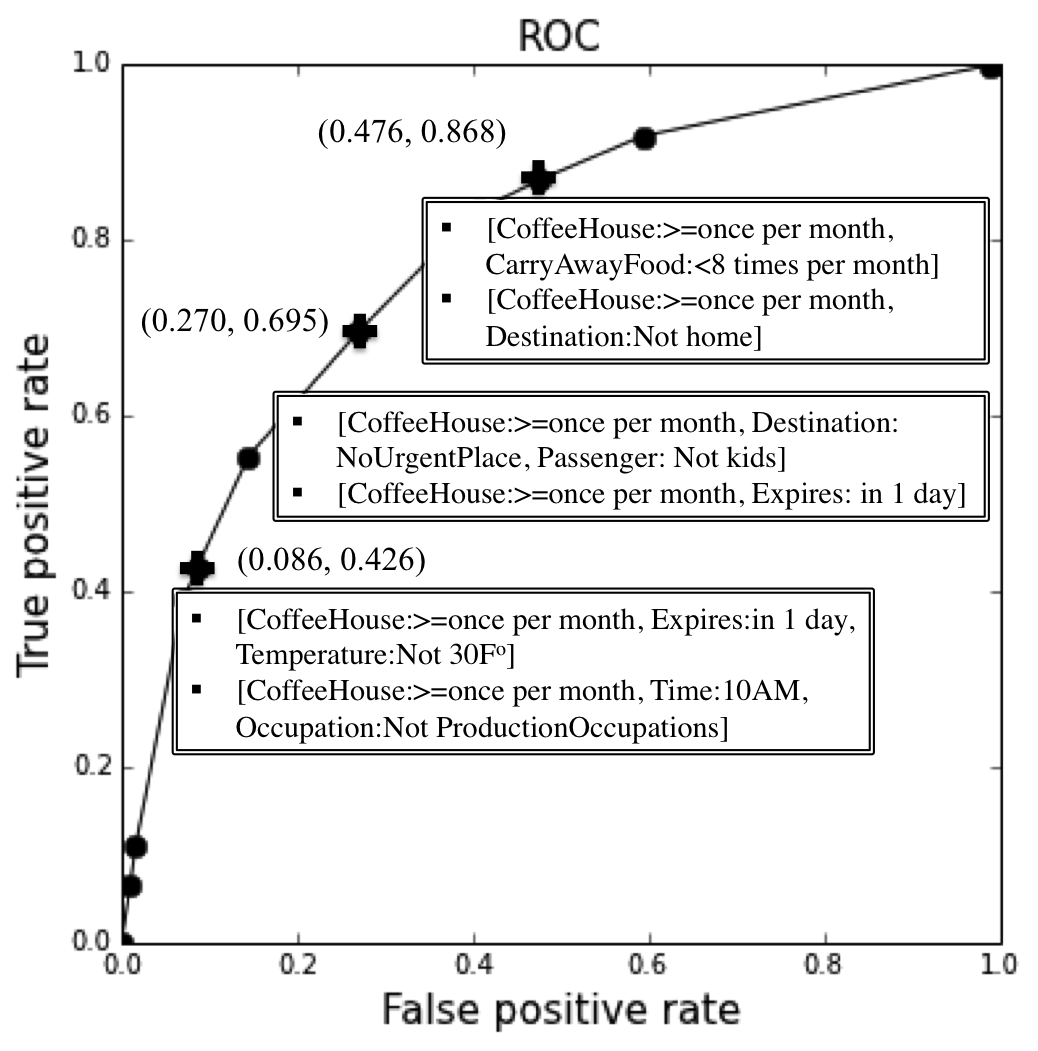}}
\hfill
\caption{ROC for dataset of coupons for bars and coffee houses}
\label{fig:roc_ex}
\end{figure}

We show several classifiers produced by BOA in Figure~\ref{fig:roc_ex}. We varied the hyperparameters $\alpha_+$,$\beta_+$,$\alpha_-$,$\beta_-$ to obtain different sets of patterns, and plotted corresponding points on the curve. Example pattern sets are listed in each box along the curve. For instance, the classifier near the middle of the curve in Figure~\ref{fig:roc_ex} (a) has one pattern, and reads ``If a person visits a bar at least once per month, is not traveling with kids, and their occupation is not farming/fishing/forestry, then predict the person will use the coupon for a bar before it expires." 
In these examples (and generally), we see that a user's general interest in a coupon's venue (bar, coffee shop, etc.) is the most relevant attribute to the classification outcome; it appears in every pattern in the two figures. 

\subsection{Experiments with UCI data sets}
We tested BOA on several datasets from the UCI machine learning repository \cite{Bache+Lichman:2013}, along with baseline algorithms, and Table~\ref{tab:UCI} displays the results. We observed that BOA achieves the best performance on each of the data sets we used. This is not a surprise: most of these data sets have an underlying true pattern set that greedy methods would have difficulty recovering. For example in the tic-tac-toe data set, the positive class can be classified using exactly 8 conditions. BOA has the capability to exactly learn these conditions, whereas the greedy splitting and pruning methods that are pervasive throughout the data mining literature (e.g., CART, C4.5) and convexified approximate methods (e.g., SVM) have substantial difficulty with this. 
We also added 30\% noise to the tic-tac-toe training set and BOA was still able to achieve perfect performance while performance of other methods suffered. Both linear models and tree models exist that achieve perfect accuracy, but the heuristic splitting/pruning and convexification of the methods we compared with prevented these perfect solutions from being found.

\begin{table*}[h]
\centering
\small
\begin{tabular}{@{}lcccccc}
\toprule
\textbf{}      & Monk 1        & Mushroom      & Breast Cancer & Connect4      & Tic-tac-toe   & \begin{tabular}[c]{@{}l@{}}Tic-tac-toe\\ (30\% noise)\end{tabular} \\ \toprule
\textbf{BOA1}            & 1.000 (0.000) & 1.000 (0.000) & 0.990 (0.003) & 0.926 (0.002) & 1.000 (0.000) & 1.000 (0.000)                                                      \\
\textbf{BOA2}           & 1.000 (0.000) & 1.000 (0.000) & 0.996 (0.007) &               0.902 (0.002) & 1.000 (0.000) & 1.000 (0.000)                                                      \\
C4.5           & 0.906 (0.067) & 1.000 (0.000) & 0.873 (0.017) & 0.867 (0.002) & 0.949 (0.016) & 0.942 (0.022)                                                      \\
CART           & 0.826 (0.061) & 1.000 (0.000) & 0.978 (0.010) & 0.703 (0.003) & 0.966 (0.011) & 0.962 (0.014)                                                      \\
RF  & 1.000 (0.000) & 1.000 (0.000) & 0.970 (0.016) & 0.940 (0.002) & 0.991 (0.003) & 0.989 (0.006)                                                      \\
Lin-Lasso   & 0.556 (0.061) & 0.995 (0.002) & 0.985 (0.005) & 0.858 (0.002) & 0.986 (0.002) & 0.854 (0.019)                                                      \\
Lin-Ridge   & 0.560 (0.078) & 0.999 (0.000) & 0.987 (0.003) & 0.857 (0.002) & 0.931 (0.017) & 0.820 (0.033)                                                      \\
Logi-Lasso & 0.666 (0.084) & 0.989 (0.002) & 0.988 (0.003) & 0.859 (0.002) & 0.988 (0.002) & 0.860 (0.029)                                                      \\
Logi-Ridge & 0.686 (0.103) & 0.999 (0.000) & 0.988 (0.003) & 0.857 (0.002) & 0.869 (0.025) & 0.805 (0.032)                                                      \\
SVM            & 0.957 (0.034) & 0.999 (0.000) & 0.986 (0.005) & 0.924 (0.002) & 0.993 (0.001) &  0.992 (0.002)  \\
\bottomrule                                                               
\end{tabular}
\caption{AUC comparison for some UCI data sets}
\label{tab:UCI}
\end{table*}

As an example, we illustrate BOA's output on the breast cancer dataset. BOA took exactly 2 minutes on a laptop to generate the following pattern set:
\begin{algorithmic}
 \If { $X$ satisfies (Marginal Adhesion $\geq 3$ AND Uniformity of Cell Shape $\geq 3$) \\  OR (Clump Thickness $\geq 7$) \\  OR (Bland Chromatin $\geq 4$ AND Uniformity of Cell Size $\geq 1$ AND Clump Thickness $\geq 2$) }
 \State Predict the tumor is malignant 
 \Else
 \State Predict the tumor is benign.
 \EndIf
\end{algorithmic}
The out-of-sample accuracy of this model was 0.952, with true positive rate 0.974, and false positive rate 0.060. \textit{Or}'s of \textit{and}'s models could potentially be useful for medical applications, since they could characterize simple sets of conditions that would place a patient in a high risk category. This may be more useful in some cases than the typical scoring systems (linear models) used in medical calculators.


\section{Conclusion} We presented a method that produces \emph{or}'s of \emph{and}'s models, where the shape of the model can be controlled by the user through Bayesian priors. In some applications, such as those arising in customer behavior modeling, the form of these models may be more useful than traditional linear models. Since finding sparse models is computationally hard, most approaches take severe heuristic approximations (such as greedy splitting and pruning in the case of decision trees, or convexification in the case of linear models). These approximations can severely hurt performance, as is easily shown experimentally, using datasets whose ground truth formulas are not difficult to find. We chose a different type of approximation, where we make an up-front statistical assumption in building our models out of pre-mined rules, and find the globally optimal solution in the reduced space of rules. We then find theoretical conditions under which using pre-mined rules provably does not change the set of MAP optimal solutions. These conditions relate the size of the dataset to the strength of the prior. If the prior is sufficiently strong and the dataset is not too large, the set of pre-mined rules is provably sufficient. We showed the benefits of this approach on a consumer behavior modeling application of current interest to ``connected vehicle" projects. Our results, using data from an extensive survey taken by several hundred individuals, show that simple patterns based on a user's context can be directly useful in predicting the user's response.

\subsection*{Acknowledgement}
We acknowledge funding from Ford Motor Company, Wistron, and Siemens.

\bibliographystyle{plain}
\bibliography{rules} 
\appendix
\section{Proofs}\label{SectionProofs}
We start with the following lemma that we will need later.
\begin{lm}
Define a function $g(l;\lambda,J)=\left(\frac{\lambda}{2}\right)^l\Gamma(J-l+1),\lambda,J \in\mathbb{N}^+$. If $1\leq l \leq J$, $g(l;\lambda,J)\leq g_\text{max}(\lambda,J)$ where $g_\text{max}(\lambda,J) = \max \left\{\left(\frac{\lambda}{2}\right)\Gamma(J),\left(\frac{\lambda}{2}\right)^J \right\}$.
\end{lm}

\begin{proof*}(Of Lemma 1)
In order to bound $g(l;\lambda,J)$, we will show that $g(l;\lambda,J)$ is convex, which means its maximum value occurs at the endpoints of the interval we are considering. The second derivative of $g(l;\lambda,J)$ respect to $l$ is
\small
\begin{equation*}
g^{\prime\prime}(l;\lambda,J) = g(l;\lambda,J)\left[\left(\ln \frac{\lambda}{2}+\gamma-\sum_{k=1}^{\infty}\frac{1}{k}-\frac{1}{k+J-L_m}\right)^2 + \sum_{k=1}^{\infty}\frac{1}{(k+J-l)^2}\right]>0,
\end{equation*}
since at least one of the terms $\frac{1}{(k+J-l)^2}>0$. Thus $g(l;\lambda,J)$ is strictly convex. Therefore the maximum of $g(l;\lambda,J)$ is achieved at the boundary of $L_m$, namely 1 or $J$. So we have
\begin{align}
g(l;\lambda,J)&\leq \max \left\{g(1;\lambda,J), g(J;\lambda,J)\right\} \notag \\
&=\max \left\{\left(\frac{\lambda}{2}\right)\Gamma(J),\left(\frac{\lambda}{2}\right)^J \right\} \notag \\
& = g_\text{max}(\lambda,J). \label{eqn:g1m}
\end{align}

\end{proof*}

\begin{proof*}(Of Theorem \ref{Theorem1})

Let $\emptyset$ denote an empty set where there are no patterns and $A^*$ is the MAP pattern set. Since $A^* \in \arg\min_A E_S(A)$, we have
\small
\begin{equation}
P(S,A^*;\theta) \geq P(S, \emptyset;\theta)\label{eqn:cp_jp},
\end{equation}
and the joint probabilities can be written as
\begin{align}
P(S,A^*;\theta) &= P(A^*;\theta)P(S|A^*;\theta)\label{eqn:a_jp},\\
P(S,\emptyset;\theta) &= P(\emptyset;\theta)P(S|\emptyset;\theta). \label{eqn:empty_jp}
\end{align}

Now we bound priors and likelihoods for these two joint probabilities. \\
\textbf{Step 1: Upper Bound for} $P(S,A^*;\theta)$.
We first look at the prior and likelihood for the MAP pattern set $A^*$. The lengths of patterns in $A^*$ are denoted as $L_m, m \in\{1,...,M^*\}$, so the prior probability of selecting $A^*$ is
\small
\begin{align}
P(A^*;\theta)& = \omega(\lambda_M,\lambda_L)\text{Poisson}(M^*;\lambda_M) \prod_m^{M^*} \text{Poisson}(L_m;\lambda_L)\frac{1}{\binom{J}{L_m}}\prod_k^{L_m}\frac{1}{K_{v_{m,k}}}\notag \\
&=\omega(\lambda_M,\lambda_L)\text{Poisson}(M^*;\lambda_M) \prod_m^{M^*}\frac{e^{-\lambda_L}\lambda_L^{L_m}\Gamma(J-L_m+1)}{\Gamma(J+1)}\prod_k^{L_m}\frac{1}{K_{v_{m,k}}} \notag \\
& \leq \omega(\lambda_M,\lambda_L)\text{Poisson}(M^*;\lambda_M) \left(\frac{e^{-\lambda_L}}{\Gamma(J+1)}\right)^{M^*}\prod_m^{M^*}\left(\frac{\lambda_L}{2}\right)^{L_m}\Gamma(J-L_m+1)\label{eqn:A1}.
\end{align}
(\ref{eqn:A1}) follows from $K_{v_{m,k}}\geq 2$ since all attributes have at least two values. Using Lemma 1 we have that 
\begin{equation}\label{eqn:g_11}
g(L_m;\lambda_L,J) = \left(\frac{\lambda_L}{2}\right)^{L_m}\Gamma(J-L_m+1) \leq g_\text{max}(\lambda,J).
\end{equation}
Combining (\ref{eqn:A1}) and (\ref{eqn:g_11}) we have 
\small
\begin{align}
P(A^*;\theta)& \leq \omega(\lambda_M,\lambda_L)\text{Poisson}(M^*;\lambda_M) \left(\frac{e^{-\lambda_L}g_\text{max}(\lambda,J)}{\Gamma(J+1)}\right)^{M^*}.
\label{eqn:best_prior}
\end{align}
The maximum likelihood of data is achieved when all points in the training set are classified correctly, and the likelihood is 1,
\small
\begin{equation} \label{eqn:a_lh}
P(S|A^*;\theta) \leq 1.
\end{equation}
Combining (\ref{eqn:best_prior}) and (\ref{eqn:a_lh}), the joint probability of $S$ and $A^*$ obeys
\small
\begin{align}\label{eqn:a_jp_bound}
P(S,A^*;\theta)& = P(A^*;\theta)P(S|A^*;\theta) \notag \\
&\leq \omega(\lambda_M,\lambda_L)\text{Poisson}(M^*;\lambda_M) \left(\frac{e^{-\lambda_L}g_\text{max}(\lambda,J)}{\Gamma(J+1)}\right)^{M^*}.
\end{align}

\noindent\textbf{Step 2: Lower Bound for} $P(S,\emptyset;\theta)$. Next we bound the prior and likelihood for an empty set. The prior probability for selecting an empty set is
\small
\begin{equation}\label{eqn:empty_prior}
P(\emptyset;\theta) = \omega(\lambda_M,\lambda_L)\text{Poisson}(0;\lambda_M) = \omega(\lambda_M,\lambda_L)e^{-\lambda_M}.
\end{equation}
Therefore the joint probability of $S$ and $\emptyset$ is:
\small
\begin{equation}\label{eqn:empty_jp_eqn}
P(S,\emptyset;\theta) = P(\emptyset;\theta)P(S|\emptyset;\theta) =  \omega(\lambda_M,\lambda_L)e^{-\lambda_M} P(S|\emptyset;\theta) .
\end{equation}

\noindent \textbf{Combine Steps 1 and 2.} Now we use inequality in (\ref{eqn:cp_jp}), and substitute individual terms with (\ref{eqn:a_jp_bound}) and (\ref{eqn:empty_jp_eqn}), we get
\small
\begin{align}
\omega(\lambda_M,\lambda_L)\frac{e^{-\lambda_M}\lambda_M^{M^*}}{M^*!} \left(\frac{e^{-\lambda_L}g_\text{max}(\lambda,J)}{\Gamma(J+1)}\right)^{M^*}&\geq  \omega(\lambda_M,\lambda_L) e^{-\lambda_M} P(S|\emptyset;\theta) \notag \\
\frac{\left(\frac{e^{-\lambda_L}g_\text{max}(\lambda,J)\lambda_M}{\Gamma(J+1)}\right)^{M^*}}{M^*!} &\geq P(S|\emptyset;\theta) \label{eqn:comb1}.
\end{align}
For simplicity we define $x = \frac{e^{-\lambda_L}g_\text{max}(\lambda,J)\lambda_M}{\Gamma(J+1)}$. If $M^* \leq \lambda_M$, the statement of the theorem holds trivially. For the remainder of the proof we consider, if $M > \lambda_M$, the left side of (\ref{eqn:comb1}) becomes $\frac{x^{M^*}}{M^*!}$ and can be upper bounded by
\small
\begin{equation*}
\frac{x^{M^*}}{M^*!} \leq \frac{x^{\lambda_M}}{\lambda_M!}\frac{x^{(M^*-\lambda_M)}}{(\lambda_M+1)^{(M^*-\lambda_M)}},
\end{equation*}
where in the denominator we used $M^*!=\lambda_M!(\lambda_M+1)\cdots M^* \geq \lambda_M!(\lambda_M+1)^{(M^*-\lambda_M)}$. So we have
\small
\begin{equation}\label{eqn:Mstar_23}
\frac{x^{\lambda_M}}{\lambda_M!}\left(\frac{x}{\lambda_M+1}\right)^{(M^*-\lambda_M)} \geq P(S|\emptyset;\theta) .
\end{equation}
By the theorem's assumption, we have $\frac{e^{-\lambda_L}\left(\frac{\lambda_L}{2}\right)^J}{\Gamma(J+1)}\leq 1$, so $\frac{e^{-\lambda_L}\left(\frac{\lambda_L}{2}\right)^J\lambda_M}{\Gamma(J+1)(\lambda_M+1)}< 1$. We also have $\frac{e^{-\lambda_L}\left(\frac{\lambda_L}{2}\right)\Gamma(J)\lambda_M}{\Gamma(J+1)(\lambda_M+1)}\leq 1$. To see this, note $\frac{\Gamma(J)}{\Gamma(J+1)}<1, \frac{\lambda_M}{\lambda_M+1}<1$ and $e^{-\lambda_L}\left(\frac{\lambda_L}{2}\right)<1$ for every $\lambda_L$. Therefore $\frac{x}{\lambda_M+1} = \frac{e^{-\lambda_L}g_\text{max}(\lambda,J)\lambda_M}{\Gamma(J+1)(\lambda_M+1)}<1$. Solving for $M^*$ in (\ref{eqn:Mstar_23}), using $\frac{x}{\lambda_M+1}<1$ to determine the direction of the inequality yields:
\begin{equation} \label{eqn:2Mstar}
M^* \leq  \lambda_M + \frac{\log P(S|\emptyset;\theta)  \cdot \frac{\lambda_M!}{x^{\lambda_M}}}{\log \frac{x}{\lambda_M + 1}} .
\end{equation}
Now we compute the likelihood for the empty set BOA model. The empty set classifies all data points as negative, so $\text{TP} = \text{FP} = 0, \text{FN} = |S^+|, \text{TN} = |S^-|$. The likelihood is 
\small
\begin{equation}\label{eqn:empty_lh}
P(S|\emptyset;\theta) = \frac{\Gamma(\alpha_-+\beta_-)}{\Gamma(\alpha_-)\Gamma(\beta_-)}\frac{\Gamma(|S^-|+\alpha_-)\Gamma(|S^+|+\beta_-)}{\Gamma(|S|+\alpha_-+\beta_-)}.
\end{equation}
Combining (\ref{eqn:2Mstar}) with (\ref{eqn:empty_lh}) and also substituting $x$ with its definition $\frac{e^{-\lambda_L}\lambda_M\max \big\{\left(\frac{\lambda_L}{2}\right)\Gamma(J),\left(\frac{\lambda_L}{2}\right)^J \big\}}{\Gamma(J+1)}$, we have
\small
\begin{equation}
M^* \leq \lambda_M + \frac{\log\left( \frac{\Gamma(\alpha_-+\beta_-)}{\Gamma(\alpha_-)\Gamma(\beta_-)}\frac{\Gamma(|S^-|+\alpha_-)\Gamma(|S^+|+\beta_-)}{\Gamma(|S|+\alpha_-+\beta_-)}\cdot\frac{\lambda_M!}{\left( \frac{e^{-\lambda_L}\lambda_M \max \big\{\left(\frac{\lambda_L}{2}\right)\Gamma(J),\left(\frac{\lambda_L}{2}\right)^J \big\}}{\Gamma(J+1)}\right)^{\lambda_M}}\right)}{\log \left(\frac{e^{-\lambda_L}\lambda_M \max \big\{\left(\frac{\lambda_L}{2}\right)\Gamma(J),\left(\frac{\lambda_L}{2}\right)^J \big\}}{\Gamma(J+1)(\lambda_M+1)}\right)}.
\end{equation}
\end{proof*}

\begin{proof*}(Of Theorem \ref{Theorem2})
Since the BOA-BetaBinomial and BOA-Poisson have the same likelihood model, the inequality (\ref {eqn:cp_jp}) still holds and we can reuse the likelihood expressions (\ref{eqn:a_lh}) and (\ref{eqn:empty_lh}).

For an empty model, the likelihood is in (\ref{eqn:empty_lh}). The BOA-BetaBinomial prior for an empty set is
\small
\begin{align}\label{eqn:empty_prior2}
P(\emptyset;\theta) &= \prod_l^L\frac{\Gamma(\alpha_l + \beta_l)}{\Gamma(\beta_l)}\frac{\Gamma(|\mathcal{A}_S^{[l]}|+\beta_l)}{\Gamma(|\mathcal{A}_S^{[l]}|+\alpha_l+\beta_l)}.
\end{align}
For the MAP solution $A^*$, we define that there are $M_l^*$ patterns selected from $\mathcal{A}_S^{[l]}$ for $l \in \{1,..,L\}$, and $M^* = \sum_l^L M_l^*.$
The prior probability of selecting $A^*$ is
\small
\begin{align}
P(A^*;\theta) &= \prod_l^L\frac{\Gamma(\alpha_l + \beta_l)}{\Gamma(\alpha_l)\Gamma(\beta_l)}\frac{\Gamma(M_l^*+\alpha_l)\Gamma(|\mathcal{A}_S^{[l]}|-M_l^*+\beta_l)}{\Gamma(|\mathcal{A}_S^{[l]}|+\alpha_l+\beta_l)} \notag \\
& = \prod_l^L\frac{\Gamma(\alpha_l + \beta_l)g_2(M_l^*)}{\Gamma(\alpha_l)\Gamma(\beta_l)\Gamma(|\mathcal{A}_S^{[l]}|+\alpha_l+\beta_l)},\label{eqn:A2}
\end{align}
where we define 
\begin{equation}\label{eqn:g1}
g_2(M_l^*) = \Gamma(M_l^*+\alpha_l)\Gamma(|\mathcal{A}_S^{[l]}|+\beta_l-M_l^*).
\end{equation}
Now we apply the same inequality
$P(A^*;\theta)P(S|A^*;\theta) \geq P(\emptyset;\theta)P(S|\emptyset;\theta)$ used for the proof for Theorem 1 and substituting in (\ref{eqn:a_lh}), (\ref{eqn:empty_lh}), (\ref{eqn:empty_prior2}) and (\ref{eqn:A2}) on individual terms, we find
\small
\begin{align}
\prod_l^L\frac{\Gamma(\alpha_l + \beta_l)g_2(M_l^*)}{\Gamma(\alpha_l)\Gamma(\beta_l)\Gamma(|\mathcal{A}_S^{[l]}|+\alpha_l+\beta_l)} &\geq P(S|\emptyset;\theta)\cdot\prod_l^L\frac{\Gamma(\alpha_l + \beta_l)}{\Gamma(\beta_l)}\frac{\Gamma(|\mathcal{A}_S^{[l]}|+\beta_l)}{\Gamma(|\mathcal{A}_S^{[l]}|+\alpha_l+\beta_l)} \notag \\
\prod_l^Lg_2(M_l^*)& \geq P(S|\emptyset;\theta)\cdot\prod_l^L\Gamma(|\mathcal{A}_S^{[l]}|+\beta_l)\Gamma(\alpha_l)\label{eqn:g2_prod} .
\end{align}
We want to find a lower bound for $g_2(M_l^*)$ for each $l$. To do this, we use 
\begin{equation}
\prod_l^L g_2(M_l^*) = g_2(M_{l^\prime}^*)\prod_{l\neq l^\prime}^Lg_2(M_l^*),
\end{equation}
where we will trivially upper bound $\prod_{l\neq l^\prime}^Lg_2(M_l^*)$ to leave only the $g_2(M_{l^\prime}^*)$ term. In particular, we observe $g_2(M_l^*) \leq g_2(0)$ for all $M_l^*$. To obtain this, we take the derivative of $g_2(M_l^*)$ with respect to $M_l^*$, as follows:
\small
\begin{align*}
g_2^\prime(M_l^*) = &g_2(M_l^*)\left(\sum_{k=1}^{\infty}\frac{1}{k+|\mathcal{A}_S^{[l]}|+\beta_l-M_l^*-1}-\frac{1}{k+M_l^*+\alpha_l-1}\right),
\end{align*}
since by the assumption of the theorem $\alpha_l<\beta_l$, thus $\alpha_l<|\mathcal{A}_S^{[l]}|+\beta_l$, and each $M_l^*\leq |\mathcal{A}_S^{[l]}|$, which means each term in the summation is less than 0. Thus $g_2^\prime(M_l^*)<0$, hence $g_2(M_l^*) \leq g_2(0)$. Therefore we can derive an inequality for each $l$:
\small
\begin{align}
\prod_l^Lg_2(M_l^*)&\leq g_2(M_{l^\prime}^*)\prod_{l=1,...L,l\neq l^\prime}g_2(0) \text{  for }l\in \{1,...,L\} \notag \\
&=g_2(M_{l^\prime}^*)\prod_{l=1,...L,l\neq l^\prime}\Gamma(\alpha_l)\Gamma(|\mathcal{A}_S^{[l]}|+\beta_l) \label{eqn:g2_prod2} .
\end{align}
Combining (\ref{eqn:g2_prod}) with (\ref{eqn:g2_prod2}) we have
\begin{equation}\label{eqn:g12}
g_2(M_l^*) \geq P(S|\emptyset;\theta)\Gamma(|\mathcal{A}_S^{[l]}|+\beta_l)\Gamma(\alpha_l) \quad\text{  for } l\in\{1,...,L\}.
\end{equation}
We write out $g_2(M_l^*)$ in (\ref{eqn:g1}) as the following, multiplying by 1 in disguise:
\small
\begin{align}
g_2(M_l^*) = &\Gamma(\alpha_l)\alpha_l\dots(\alpha_l+M_l^*-1)\frac{\Gamma(|\mathcal{A}_S^{[l]}|+\beta_l-M_l^*)(|\mathcal{A}_S^{[l]}|+\beta_l-M_l^*)\dots(|\mathcal{A}_S^{[l]}|+\beta_l-1)}{(|\mathcal{A}_S^{[l]}|+\beta_l-M_l^*)\dots(|\mathcal{A}_S^{[l]}|+\beta_l-1)}\notag \\
=&\Gamma(\alpha_l)\Gamma(|\mathcal{A}_S^{[l]}|+\beta_l)\frac{\alpha_l\dots(\alpha_l+M_l^*-1)}{(|\mathcal{A}_S^{[l]}|+\beta_l-M_l^*)\dots(|\mathcal{A}_S^{[l]}|+\beta_l-1)}\notag \\
\leq &\Gamma(\alpha_l)\Gamma(|\mathcal{A}_S^{[l]}|+\beta_l)\left(\frac{\alpha_l+M_l^*-1}{|\mathcal{A}_S^{[l]}|+\beta_l-1}\right)^{M_l^*} \notag \\
\leq &\Gamma(\alpha_l)\Gamma(|\mathcal{A}_S^{[l]}|+\beta_l)\left(\frac{|\mathcal{A}_S^{[l]}|+\alpha_l-1}{|\mathcal{A}_S^{[l]}|+\beta_l-1}\right)^{M_l^*} \label{eqn:g1_last}.
\end{align}
Here, (\ref{eqn:g1_last}) follows because $M_l^* \leq |\mathcal{A}_S^{[l]}|$. Combining with (\ref{eqn:g12}), we have
\begin{equation*}
P(S|\emptyset;\theta) \leq \left(\frac{|\mathcal{A}_S^{[l]}|+\alpha_l-1}{|\mathcal{A}_S^{[l]}|+\beta_l-1}\right)^{M_l^*}.
\end{equation*}
Substituting in equation (\ref{eqn:empty_lh}) and using $\alpha_l<\beta_l$, we get
\begin{equation}\label{eqn:Ml}
 M_l^* \leq \frac{\log  \frac{\Gamma(\alpha_-+\beta_-)}{\Gamma(\alpha_-)\Gamma(\beta_-)}\frac{\Gamma(|S^-|+\alpha_-)\Gamma(|S^+|+\beta_-)}{\Gamma(|S|+\alpha_-+\beta_-)}}{\log \frac{|\mathcal{A}_S^{[l]}|+\alpha_l-1}{|\mathcal{A}_S^{[l]}|+\beta_l-1}},
\end{equation}
which holds for $l \in\{1...L\}$. Thus
\begin{equation} 
M^* = \sum_l M_l^* \leq \sum_l^L\frac{\log  \frac{\Gamma(\alpha_-+\beta_-)}{\Gamma(\alpha_-)\Gamma(\beta_-)}\frac{\Gamma(|S^-|+\alpha_-)\Gamma(|S^+|+\beta_-)}{\Gamma(|S|+\alpha_-+\beta_-)}}{\log \frac{|\mathcal{A}_S^{[l]}|+\alpha_l-1}{|\mathcal{A}_S^{[l]}|+\beta_l-1}}.
\end{equation}
\end{proof*}

\begin{proof*}(Of Theorem \ref{Theorem3})
For a pattern set $A$, we will show that if any pattern $a_z$ has support $\text{supp}_S(a_z)<C$ on data $S$, then $A \not\in \underset{A^\prime\in\Lambda_S}{\arg\min}E_S(A^\prime)$. Assume pattern $a_z$ has support less than C and define $A_{\backslash z}$ which has the $z$-th pattern removed from $A$: $$A_{\backslash z}=\{a_i\}_{i=1,...,|A|, i\neq z}.$$ Assume $A$ consists of $M$ rules, among which $M_l$ come from pool $\mathcal{A}_S^{[l]}$ of rules with length $l$, $l \in \{1,...L\}$, and the $z$-th rule has length $l^\prime$ so it is removed from $\mathcal{A}_S^{[l^\prime]}$. Define TP, FP, TN and FN to be the number of true positives, false positives, true negatives and false negatives in $S$ given $A$.
We now compute the likelihood for model $A_{\backslash z}$. The most extreme case is when pattern $a_z$ is an accurate rule that applies only to real positive data points and those data points satisfy only $a_z$. Therefore once removing it, the number of true positives decreases by $\text{supp}_S(a_z)$ and the number of false negatives increases by $\text{supp}_S(a_z)$.

\noindent\textbf{Step 1: Relate $P(S|A_{\backslash z};\theta)$ to $P(S|A;\theta)$. }
\begin{align}\label{eqn:Aj_lh}
P(S|A_{\backslash z},\theta) \geq & \frac{\Gamma(\alpha_+ + \beta_+)}{\Gamma(\alpha_+)\Gamma(\beta_+)}\frac{\Gamma(\text{TP}+\alpha_+-\text{supp}_S(a_z))\Gamma(\text{FP}+\beta_+)}{\Gamma(\text{TP}+\text{FP}+\alpha_++\beta_+-\text{supp}_S(a_z))} \notag \\
& \cdot \frac{\Gamma(\alpha_- + \beta_-)}{\Gamma(\alpha_-)\Gamma(\beta_-)}\frac{\Gamma(\text{TN}+\alpha_-)\Gamma(\text{FN}+\beta_-+\text{supp}_S(a_z))}{\Gamma(\text{TN}+\text{FN}+\alpha_-+\beta_-+\text{supp}_S(a_z))} \notag \\
=&P(S|A;\theta)\cdot g_3(\text{TP},\text{FP},\text{TN},\text{FN};\alpha_+,\beta_+,\alpha_-,\beta_-,\text{supp}_S(a_z)),
\end{align}
where
\small
\begin{align}\label{eqn:g3}
g_3(\text{TP},\text{FP},\text{TN},\text{FN};\alpha_+,\beta_+,&\alpha_-,\beta_-,\text{supp}_S(a_z))= \notag \\
&\frac{\Gamma(\text{TP}+\alpha_+-\text{supp}_S(a_z))}{\Gamma(\text{TP}+\alpha_+)}\frac{\Gamma(\text{TP}+\text{FP}+\alpha_++\beta_+)}{\Gamma(\text{TP}+\text{FP}+\alpha_++\beta_+-\text{supp}_S(a_z))} \cdot \notag \\
&\frac{\Gamma(\text{FN}+\beta_-+\text{supp}_S(a_z))}{\Gamma(\text{FN}+\beta_-)} \frac{\Gamma(\text{TN}+\text{FN}+\alpha_-+\beta_-)}{\Gamma(\text{TN}+\text{FN}+\alpha_-+\beta_-+\text{supp}_S(a_z))}.
\end{align}
Now we break down $g_3(\text{TP},\text{FP},\text{TN},\text{FN};\alpha_+,\beta_+,\alpha_-,\beta_-,\text{supp}_S(a_z))$ to find a lower bound for it. The first two terms in (\ref{eqn:g3}) become
\small
\begin{align}\label{eqn:g3_1}
\frac{\Gamma(\text{TP}+\alpha_+-\text{supp}_S(a_z))}{\Gamma(\text{TP}+\alpha_+)}&\frac{\Gamma(\text{TP}+\text{FP}+\alpha_++\beta_+)}{\Gamma(\text{TP}+\text{FP}+\alpha_++\beta_+-\text{supp}_S(a_z))}\notag \\
&= \frac{(\text{TP}+\text{FP}+\alpha_++\beta_+-\text{supp}_S(a_z))\dots(\text{TP}+\text{FP}+\alpha_++\beta_+-1)}{(\text{TP}+\alpha_+-\text{supp}_S(a_z))\dots (\text{TP}+\alpha_+-1)} \notag \\
& \geq \left(\frac{\text{TP}+\text{FP}+\alpha_++\beta_+-1}{\text{TP}+\alpha_+-1} \right)^{\text{supp}_S(a_z)} \notag \\
& \geq \left(\frac{|S^+|+\alpha_++\beta_+-1}{|S^+|+\alpha_+-1} \right)^{\text{supp}_S(a_z)}.
\end{align}
Equality holds in (\ref{eqn:g3_1}) when $\text{TP} = |S^+|, \text{FP} = 0$.
The last two terms in (\ref{eqn:g3}) become
\small
\begin{align}
\frac{\Gamma(\text{FN}+\beta_-+\text{supp}_S(a_z))}{\Gamma(\text{FN}+\beta_-)} &\frac{\Gamma(\text{TN}+\text{FN}+\alpha_-+\beta_-)}{\Gamma(\text{TN}+\text{FN}+\alpha_-+\beta_-+\text{supp}_S(a_z))} \notag \\
&= \frac{(\text{FN}+\beta_-)\dots(\text{FN}+\beta_-+\text{supp}_S(a_z)-1)}{(\text{TN}+\text{FN}+\alpha_-+\beta_-)\dots(\text{TN}+\text{FN}+\alpha_-+\beta_-+\text{supp}_S(a_z)-1)} \notag \\
&\geq  \left(\frac{\text{FN}+\beta_-}{\text{FN}+\text{TN}+\alpha_-+\beta_-}\right)^{\text{supp}_S(a_z)}\notag \\
&\geq \left(\frac{\beta_-}{|S^-|+\alpha_-+\beta_-}\right)^{\text{supp}_S(a_z)}. \label{eqn:FN0}
\end{align}
Equality in (\ref{eqn:FN0}) holds when $\text{TN}=|S^-|, \text{FN}= 0$.
Combining (\ref{eqn:g3}), (\ref{eqn:g3_1}) and (\ref{eqn:FN0}), we obtain a lower bound for $g_3(\text{TP},\text{FP},\text{TN},\text{FN};\alpha_+,\beta_+,\alpha_-,\beta_-,\text{supp}_S(a_z))$ as
\small
\begin{equation*}
g_3(\text{TP},\text{FP},\text{TN},\text{FN},\alpha_+,\beta_+,\alpha_-,\beta_-,\text{supp}_S(a_z)) \geq \left(\frac{|S^+|+\alpha_++\beta_+-1}{|S^+|+\alpha_+-1}\frac{\beta_-}{|S^-|+\alpha_-+\beta_-}\right)^{\text{supp}_S(a_z)}. 
\end{equation*}
Following (\ref{eqn:Aj_lh}),
\small
\begin{equation}\label{eqn:Aj_lh_final}
P(S|A_{\backslash z},\theta) \geq\left(\frac{|S^+|+\alpha_++\beta_+-1}{|S^+|+\alpha_+-1}\frac{\beta_-}{|S^-|+\alpha_-+\beta_-}\right)^{\text{supp}_S(a_z)} \cdot P(S|A;\theta).
\end{equation}
\textbf{Step 2: Relate $P(A_{\backslash z};\theta)$ to $P(A;\theta)$.}
Since $A_{\backslash z}$ consists of the same rules as $A$ except missing one rule from $\mathcal{A}_S^{[l^\prime]}$, we multiply $P(A_{\backslash z};\theta)$ with 1 in disguise to relate it to $P(A;\theta)$.
\small
\begin{align*} 
P(A_{\backslash z};\theta) &=\frac{\Gamma(\alpha_{l^\prime} + \beta_{l^\prime})}{\Gamma(\alpha_{l^\prime})\Gamma(\beta_{l^\prime})}\frac{\Gamma(M_{l^\prime}-1+\alpha_{l^\prime})\Gamma(|\mathcal{A}_S^{[l]}|-M_l+1+\beta_{l^\prime})}{\Gamma(|\mathcal{A}_S^{[l]}|+\alpha_{l^\prime}+\beta_{l^\prime})}   \cdot \notag \\
&\quad\quad\quad\quad\quad\quad\quad\quad \prod_{l\neq l^\prime}^L\frac{\Gamma(\alpha_l + \beta_l)}{\Gamma(\alpha_l)\Gamma(\beta_l)}\frac{\Gamma(M_l+\alpha_l)\Gamma(|\mathcal{A}_S^{[l]}|-M_l+\beta_l)}{\Gamma(|\mathcal{A}_S^{[l]}|+\alpha_l+\beta_l)} \notag \\ 
&=\frac{\frac{\Gamma(\alpha_{l^\prime} + \beta_{l^\prime})}{\Gamma(\alpha_{l^\prime})\Gamma(\beta_{l^\prime})}\frac{\Gamma(M_{l^\prime}-1+\alpha_{l^\prime})\Gamma(|\mathcal{A}_S^{[l]}|-M_l+1+\beta_{l^\prime})}{\Gamma(|\mathcal{A}_S^{[l]}|+\alpha_{l^\prime}+\beta_{l^\prime})}   \cdot   \prod_{l\neq l^\prime}^L\frac{\Gamma(\alpha_l + \beta_l)}{\Gamma(\alpha_l)\Gamma(\beta_l)}\frac{\Gamma(M_l+\alpha_l)\Gamma(|\mathcal{A}_S^{[l]}|-M_l+\beta_l)}{\Gamma(|\mathcal{A}_S^{[l]}|+\alpha_l+\beta_l)}}{\prod_{l}^L\frac{\Gamma(\alpha_l + \beta_l)}{\Gamma(\alpha_l)\Gamma(\beta_l)}\frac{\Gamma(M_l+\alpha_l)\Gamma(|\mathcal{A}_S^{[l]}|-M_l+\beta_l)}{\Gamma(|\mathcal{A}_S^{[l]}|+\alpha_l+\beta_l)}}\cdot P(A;\theta) \notag \\
&=\frac{\Gamma(M_{l^\prime}-1+\alpha_{l^\prime})\Gamma(|\mathcal{A}_S^{[l^\prime]}|-M_{l^\prime}+1+\beta_{l^\prime})}{\Gamma(M_{l^\prime}+\alpha_{l^\prime})\Gamma(|\mathcal{A}_S^{[l^\prime]}|-M_{l^\prime}+\beta_{l^\prime})} \cdot  P(A;\theta) \notag \\
&=\frac{|\mathcal{A}_S^{[l^\prime]}|-M_{l^\prime}+\beta_{l^\prime}}{M_{l^\prime}-1+\alpha_{l^\prime}}\cdot P(A;\theta).
\end{align*}
$\frac{|\mathcal{A}_S^{[l^\prime]}|-M_{l^\prime}+\beta_{l^\prime}}{M_{l^\prime}-1+\alpha_{l^\prime}}$ decreases monotonically as $M_{l^\prime}$ increases, therefore it is lower bounded at the maximum of $M_{l^\prime}$. We use $m_l$ to denote the upper bound for $M_{l^\prime}$ from (\ref{eqn:Ml}), i.e.,
\small
\begin{equation*}
m_l = \frac{\log  \frac{\Gamma(\alpha_-+\beta_-)}{\Gamma(\alpha_-)\Gamma(\beta_-)}\frac{\Gamma(|S^-|+\alpha_-)\Gamma(|S^+|+\beta_-)}{\Gamma(|S|+\alpha_-+\beta_-)}}{\log \frac{|\mathcal{A}_S^{[l]}|+\alpha_l-1}{|\mathcal{A}_S^{[l]}|+\beta_l-1}},
\end{equation*}
so 
\small
\begin{equation*}
\frac{|\mathcal{A}_S^{[l^\prime]}|-M_{l^\prime}+\beta_{l^\prime}}{M_{l^\prime}-1+\alpha_{l^\prime}} \geq \frac{|\mathcal{A}_S^{[l^\prime]}|-m_{l^\prime}+\beta_{l^\prime}}{m_{l^\prime}-1+\alpha_{l^\prime}} \text{  for  }l^\prime \in \{1,...L\}.
\end{equation*}
Therefore
\small
\begin{equation}\label{eqn:Aj_final}
P(A_{\backslash z};\theta)\geq \underset{l}{\min}\left(\frac{|\mathcal{A}_S^{[l]}|-m_{l}+\beta_{l}}{m_{l}-1+\alpha_{l}}\right) P(A;\theta).
\end{equation}
\textbf{Step 3: Combine Step 1 and Step 2.}
Combining (\ref{eqn:Aj_lh_final}) and (\ref{eqn:Aj_final}), the joint probability of $S$ and $A_{\backslash z}$ is bounded by
\small
\begin{align*}
P(S,A_{\backslash z};\theta) &= P(A_{\backslash z};\theta) P(S|A_{\backslash z},\theta) \notag \\
& \geq \underset{l}{\min}\left(\frac{|\mathcal{A}_S^{[l]}|-m_{l}+\beta_{l}}{m_{l}-1+\alpha_{l}}\right)\left(\frac{|S^+|+\alpha_++\beta_+-1}{|S^+|+\alpha_+-1}\frac{\beta_-}{|S^-|+\alpha_-+\beta_-}\right)^{\text{supp}_S(a_z)}\cdot P(S,A;\theta).
\end{align*}
In order to get $P(S,A_{\backslash z};\theta) \geq P(S,A;\theta)$, we need 
\small
\begin{equation*}
\underset{l}{\min}\left(\frac{|\mathcal{A}_S^{[l]}|-m_{l}+\beta_{l}}{m_{l}-1+\alpha_{l}}\right)\left(\frac{|S^+|+\alpha_++\beta_+-1}{|S^+|+\alpha_+-1}\frac{\beta_-}{|S^-|+\alpha_-+\beta_-}\right)^{\text{supp}_S(a_z)}\geq 1.
\end{equation*}
We have $\frac{|S^+|+\alpha_++\beta_+-1}{|S^+|+\alpha_+-1}\frac{\beta_-}{|S^-|+\alpha_-+\beta_-}\leq1$ from the assumption in the theorem's statement, thus 
\small
\begin{align*}
\text{supp}_S(a_z)& \leq \frac{\log\underset{l}{\min}\left(\frac{|\mathcal{A}_S^{[l]}|-m_{l}+\beta_{l}}{m_{l}-1+\alpha_{l}}\right)}{\log \frac{|S^+|+\alpha_+-1}{|S^+|+\alpha_++\beta_+-1}\frac{|S^-|+\alpha_-+\beta_-}{\beta_-}}.
\end{align*}
\end{proof*}
\begin{proof*}(Of Theorem \ref{Theorem4})
Similar to the proof for Theorem \ref{Theorem3}, we will show that for a pattern set $A$, if any pattern $a_z$ has support $\text{supp}_S(a_z)<C$ on data $S$, then $A \not\in \underset{A^\prime\in\Lambda_S}{\arg\min}E_S(A^\prime)$. Assume pattern $a_z$ has support less than C and $A_{\backslash z}$ has the $k$-th pattern removed from $A$. Assume $A$ consists of $M$ rules, and the $z$-th rule has length $L_z$. 

Step 1 is the same as in the proof for Theorem \ref{Theorem3}, we relate $P(A_{\backslash z};\theta)$ with $P(A;\theta)$. We multiply $P(A_{\backslash z};\theta)$ with 1 in disguise to relate it to $P(A;\theta)$:
\begin{align}
P(A_{\backslash z};\theta) &=\omega(\lambda_M,\lambda_L)\text{Poisson}(M-1;\lambda_M) \prod_{m\neq z}^M \text{Poisson}(L_m;\lambda_L)\frac{1}{\binom{J}{L_m}}\prod_k^{L_m}\frac{1}{K_{v_{m,k}}} \notag \\
&=\frac{\omega(\lambda_M,\lambda_L)\text{Poisson}(M-1;\lambda_M) \prod_{m\neq z}^M \text{Poisson}(L_m;\lambda_L)\frac{1}{\binom{J}{L_m}}\prod_k^{L_m}\frac{1}{K_{v_{m,k}}}}{\omega(\lambda_M,\lambda_L)\text{Poisson}(M;\lambda_M) \prod_m^M \text{Poisson}(L_m;\lambda_L)\frac{1}{\binom{J}{L_m}}\prod_k^{L_m}\frac{1}{K_{v_{m,k}}}} P(A;\theta)\notag \\
& = \frac{M\Gamma(J+1)}{\lambda_M e^{-\lambda_L}\lambda_L^{L_z}\Gamma(J-L_z+1)\prod_k^{L_z}\frac{1}{K_{v_{z,k}}}}  P(A;\theta)  \notag \\
& \geq \frac{M\Gamma(J+1)}{\lambda_M e^{-\lambda_L}\left(\frac{\lambda_L}{2}\right)^{L_z}\Gamma(J-L_z+1)}  P(A;\theta) \label{eqn:th4_a}\\
&= \frac{M\Gamma(J+1)}{\lambda_M e^{-\lambda_L}g(L_z;\lambda_L,J)}  P(A;\theta)  \label{eqn:th4_b}\\
&\geq \frac{\Gamma(J+1)}{\lambda_M e^{-\lambda_L}g_\text{max}(\lambda_L,J)}  P(A;\theta),  \label{eqn:th4_c}
\end{align}
where (\ref{eqn:th4_a}) follows that $K_{v_{m,k}}\geq 2$ since all attributes have at least two values, (\ref{eqn:th4_b}) follows the definition of $g(l;\lambda,J)$ in Lemma 1, and (\ref{eqn:th4_c}) uses the upper bound in Lemma 1 and $M\geq 1$.

Then combining (\ref{eqn:Aj_lh_final}) with (\ref{eqn:th4_c}), the joint probability of $S$ and $A_{\backslash z}$ is lower bounded by
\small
\begin{align*}
P(S,A_{\backslash z};\theta) &= P(A_{\backslash z};\theta) P(S|A_{\backslash z},\theta) \notag \\
&\geq\frac{\Gamma(J+1)}{\lambda_M e^{-\lambda_L}g_\text{max}(\lambda_L,J)}\left(\frac{|S^+|+\alpha_++\beta_+-1}{|S^+|+\alpha_+-1}\frac{\beta_-}{|S^-|+\alpha_-+\beta_-}\right)^{\text{supp}_S(a_z)}\cdot P(S,A;\theta).
\end{align*}
In order to get $P(S,A_{\backslash z};\theta) \geq P(S,A;\theta)$, we need 
\begin{equation*}
\frac{\Gamma(J+1)}{\lambda_M e^{-\lambda_L}g_\text{max}(\lambda_L,J)}\left(\frac{|S^+|+\alpha_++\beta_+-1}{|S^+|+\alpha_+-1}\frac{\beta_-}{|S^-|+\alpha_-+\beta_-}\right)^{\text{supp}_S(a_z)}\geq 1
\end{equation*}
We have $\frac{|S^+|+\alpha_++\beta_+-1}{|S^+|+\alpha_+-1}\frac{\beta_-}{|S^-|+\alpha_-+\beta_-}\leq1$ and $g_\text{max}(\lambda_L,J) = \max \left\{\left(\frac{\lambda}{2}\right)\Gamma(J),\left(\frac{\lambda}{2}\right)^J \right\}$, thus 
\begin{equation}
\text{supp}_S(a_z) \leq \frac{\log \left(\frac{\Gamma(J+1)}{\lambda_M e^{-\lambda_L}\max \left\{\left(\frac{\lambda}{2}\right)\Gamma(J),\left(\frac{\lambda}{2}\right)^J \right\}} \right)}{\log \frac{|S^+|+\alpha_+-1}{|S^+|+\alpha_++\beta_+-1}\frac{|S^-|+\alpha_-+\beta_-}{\beta_-}}.
\end{equation}
\end{proof*}
\begin{proof*}(Of Theorem \ref{Theorem5})
Consider the empirical risk on data $S$ given $\rho_+, \rho_-$:
\begin{align}
R^{\text{emp}}(A) = &\frac{1}{N}\sum_{n=1}^{N}\mathbbm{1}_{y_n \neq f_{A}(\mathbf{x}_n)} \notag \\
  =& \frac{1}{N} \left(\sum_{f(\mathbf{x}_n)=1,y_n=1}0 +  \sum_{f(\mathbf{x}_n)=0,y_n=0}0 + \sum_{f(\mathbf{x}_n)=1,y_n=0}1  + \sum_{f(\mathbf{x}_n)=0,y_n=1}1  \right) \notag \\
 \leq &  \frac{1}{N} \Bigg(\sum_{f(\mathbf{x}_n)=1,y_n=1} \frac{\log \rho_+}{\log \frac{1}{2}} +  \sum_{f(\mathbf{x}_n)=0,y_n=0} \frac{\log \rho_-}{\log \frac{1}{2}}   \notag  \\
 & + \sum_{f(\mathbf{x}_n)=1,y_n=0} \frac{\log (1-\rho_+)}{\log \frac{1}{2}} + \sum_{f(\mathbf{x}_n)=0,y_n=1} \frac{\log (1-\rho_-)}{\log \frac{1}{2}}   \Bigg)   \label{eqn:Remp1} \\
 \leq & \frac{\log P(S|\rho_+,\rho_-)}{N\log\frac{1}{2}}.  \label{eqn:Remp2}
\end{align}
Since $\rho_+,\rho_->\frac{1}{2}$, (\ref{eqn:Remp1}) follows 
\small
\begin{align}
0 \leq  \left\{\frac{\log \rho_+}{\log \frac{1}{2}},  \frac{\log \rho_-}{\log \frac{1}{2}} \right\}  & \leq 1  \notag \\
\left\{  \frac{\log (1-\rho_+)}{\log \frac{1}{2}},  \frac{\log (1-\rho_-)}{\log \frac{1}{2}} \right\} & \geq 1 \notag
\end{align}
Using Hoeffding's Inequality and the union bound, we can get, with probability $1-\delta$, for all $A \in \Lambda^\mathcal{F}$,
\small
\begin{align}\label{eqn:Rtrue}
R^{\text{true}}(A) \leq &R^{\text{emp}}(A)+  \sqrt{\frac{|\Lambda^{\mathcal{F}}|+\log \frac{1}{\delta}}{2N}}, 
\end{align}
where $\Lambda^{\mathcal{F}}$ denotes the class of BOA models. $|\Lambda^{\mathcal{F}}|$ can be computed by counting the number of patterns sets of different sizes up to $M_{\text{upper}}$, which is in Theorem 1 and Theorem 2 for the two BOA models. 
\small
\begin{equation*}
|\Lambda^{\mathcal{F}}| = \sum_{m=1}^{M_{\text{upper}}}{|\mathcal{A}_S| \choose m}.
\end{equation*} 
$\mathcal{A}_S$ contains all patterns.
\small
\begin{equation*}
|\mathcal{A}_S| = \prod_j^{J}(K_j +1).
\end{equation*}
Therefore 
\small
\begin{equation}\label{eqn:F}
|\Lambda^{\mathcal{F}}| \leq \sum_{m=1}^{M_{\text{upper}}}{\prod_j^{J}(K_j +1) \choose m}.
\end{equation} 
Combining (\ref{eqn:Remp2}), (\ref{eqn:Rtrue}) and (\ref{eqn:F}), we have
\small
\begin{equation}
R^{\text{true}}(A) \leq   \frac{\log P(S|A;\rho_+,\rho_-)}{N\log\frac{1}{2}}  + \sqrt{\frac{\sum_{m=1}^{M_{\text{upper}}}{\prod_j^{J}(K_j +1) \choose m}+\log \frac{1}{\delta}}{2N}}.
\end{equation}
\end{proof*}

\section{Mobile Advertisement Datasets}
\begin{figure*}[h!]
\centering
\includegraphics[width=0.85\textwidth]{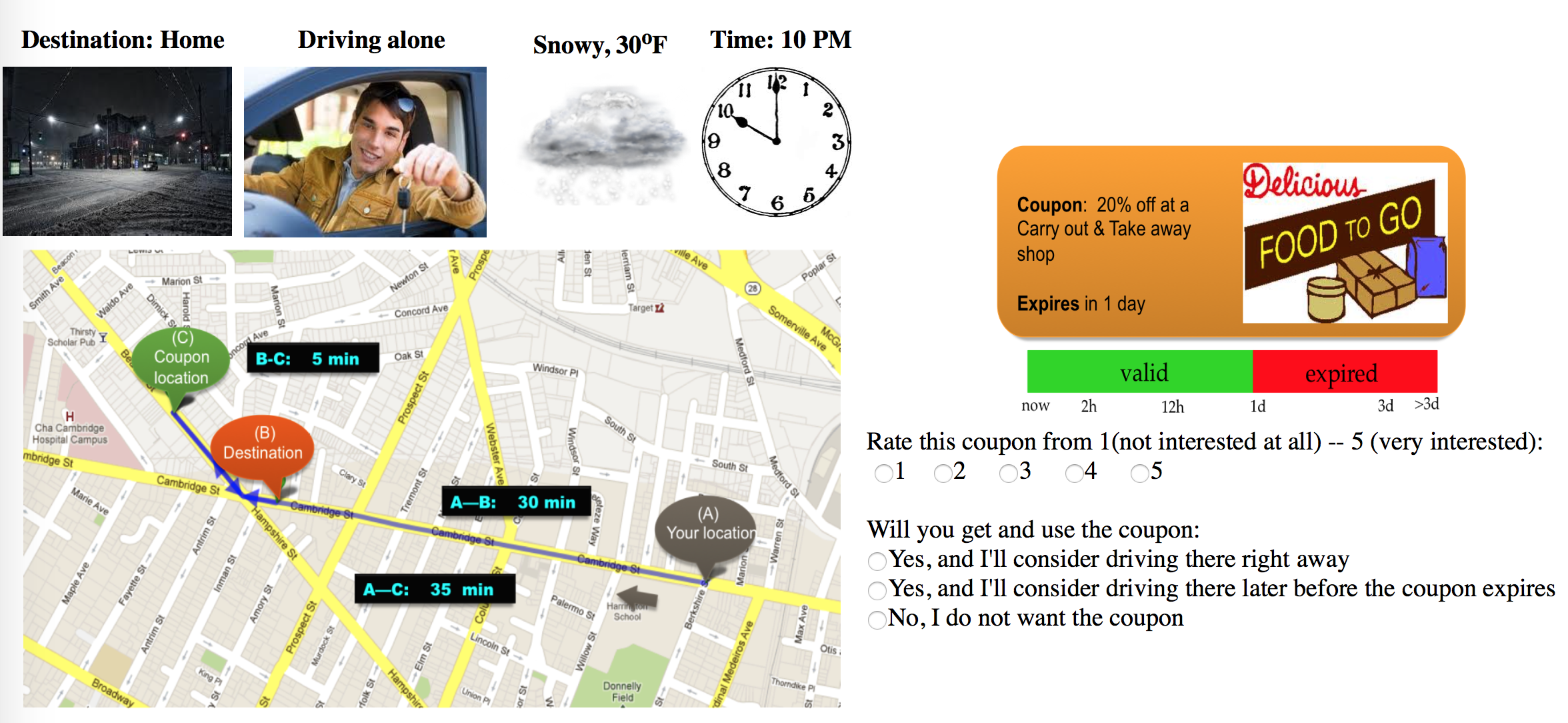}
\caption{An example of scenario in the survey}
\label{fig:survey}
\end{figure*}

The attributes of this data set include:
\begin{enumerate}
\item User attributes
\begin{itemize}
\compresslist
\item Gender: male, female
\item Age: below 21, 21 to 25, 26 to 30, etc.
\item Marital Status: single, married partner, unmarried partner, or widowed
\item Number of children: 0, 1, or more than 1
\item Education: high school, bachelors degree, associates degree, or graduate degree
\item Occupation: architecture \& engineering, business \& financial, etc.
\item Annual income: less than \$12500, \$12500 - \$24999, \$25000 - \$37499, etc.
\item Number of times that he/she goes to a bar: 0, less than 1, 1-3, 4-8 or greater than 8
\item Number of times that he/she buys takeaway food: 0, less than 1, 1-3, 4-8 or greater than 8
\item Number of times that he/she goes to a coffee house: 0, less than 1, 1-3, 4-8 or greater than 8
\item Number of times that he/she eats at a restaurant with average expense less than \$20 per person: 0, less than 1, 1-3, 4-8 or greater than 8
\item Number of times that he/she goes to a bar: 0, less than 1, 1-3, 4-8 or greater than 8
\end{itemize}
\item Contextual attributes
\begin{itemize}
\compresslist
\item Driving destination: home, work, or no urgent destination
\item Location of user, coupon and destination: we provide a map to show the geographical location of the user, destination, and the venue, and we mark the distance between each two places with time of driving. The user can see whether the venue is in the same direction as the destination.
\item Weather: sunny, rainy, or snowy
\item Temperature: $30\text{F}^\text{o}$, $55\text{F}^\text{o}$, or $80\text{F}^\text{o}$
\item Time: 10AM, 2PM, or 6PM
\item Passenger: alone, partner, kid(s), or friend(s)
\end{itemize}
\item Coupon attributes
\begin{itemize}
\compresslist
\item time before it expires: 2 hours or one day
\end{itemize}
\end{enumerate}
All coupons provide a 20\% discount.
The survey was divided into different parts, so that Turkers without children would never see a scenario where their ``kids" were in the vehicle. 
Figure ~\ref{fig:survey} shows an example of scenarios in the survey.


\end{document}